\documentclass[10pt,twocolumn,letterpaper]{article}

\usepackage{iccv}              

\definecolor{iccvblue}{rgb}{0.21,0.49,0.74}
\newcommand{\ul}[1]{\underline{#1}}

\usepackage[pagebackref,breaklinks,colorlinks,allcolors=iccvblue]{hyperref}
\usepackage{adjustbox}
\usepackage{multirow}
\usepackage{xcolor}    
\usepackage{colortbl}  
\usepackage{float}
\usepackage{stfloats}

\def\paperID{14485} 
\def\confName{ICCV}
\def\confYear{2025}

\title{INS-MMBench: A Comprehensive Benchmark for Evaluating LVLMs' Performance in Insurance}

\author{%
  Chenwei~Lin \\
  School of Computer Science\\
  Fudan University\\
  Shanghai, 200433, China \\
  {\tt\small cwlin23@m.fudan.edu.cn}
  \and
  Hanjia~Lyu \\
  Department of Computer Science\\
  University of Rochester \\
  Rochester, NY 14627, USA \\
  {\tt\small hlyu5@ur.rochester.edu}
  \and
  Xian~Xu \\
  School of Economics\\
  Fudan University \\
  Shanghai, 200433, China \\
  {\tt\small xianxu@fudan.edu.cn}
  \and
  Jiebo~Luo \\
  Department of Computer Science\\
  University of Rochester \\
  Rochester, NY 14627, USA \\
  {\tt\small jluo@cs.rochester.edu}
}

\begin{document}
\maketitle

\begin{abstract}
Large Vision-Language Models (LVLMs) and Multimodal Large Language Models (MLLMs) have demonstrated outstanding performance in various general multimodal applications and have shown increasing promise in specialized domains. However, their potential in the insurance domain—characterized by diverse application scenarios and rich multimodal data—remains largely underexplored. To date, there is no systematic review of multimodal tasks, nor a benchmark specifically designed to assess the capabilities of LVLMs in insurance. This gap hinders the development of LVLMs within the insurance industry. This study systematically reviews and categorizes multimodal tasks for 4 representative types of insurance: auto, property, health, and agricultural. We introduce \textbf{INS-MMBench}, the first hierarchical benchmark tailored for the insurance domain. INS-MMBench encompasses 22 fundamental tasks, 12 meta-tasks and 5 scenario tasks, enabling a comprehensive and progressive assessment from basic capabilities to real-world use cases. We benchmark 11 leading LVLMs, including closed-source models such as GPT-4o and open-source models like LLaVA. Our evaluation validates the effectiveness of INS-MMBench and offers detailed insights into the strengths and limitations of current LVLMs on a variety of insurance-related multimodal tasks. We hope that INS-MMBench will accelerate the integration of LVLMs into the insurance industry and foster interdisciplinary research. Our dataset and evaluation code are  available at \url{https://github.com/FDU-INS/INS-MMBench}.
\end{abstract}    
\begin{figure*}
    \centering
    \begin{subfigure}{0.43\textwidth}
        \centering
        \includegraphics[width=\textwidth]{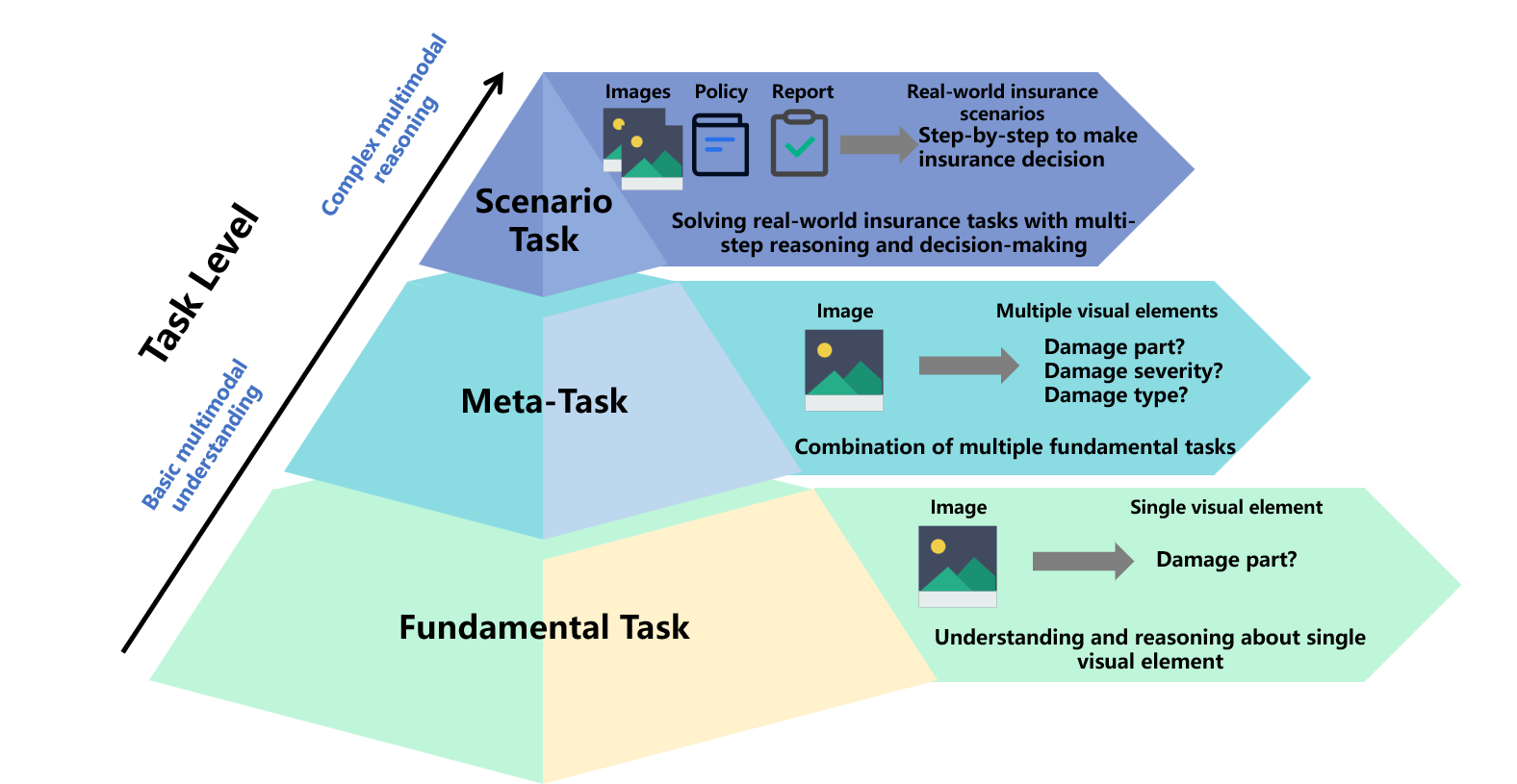}
    \end{subfigure}
    \hfill
    \begin{subfigure}{0.43\textwidth}
        \centering
        \includegraphics[width=\textwidth]{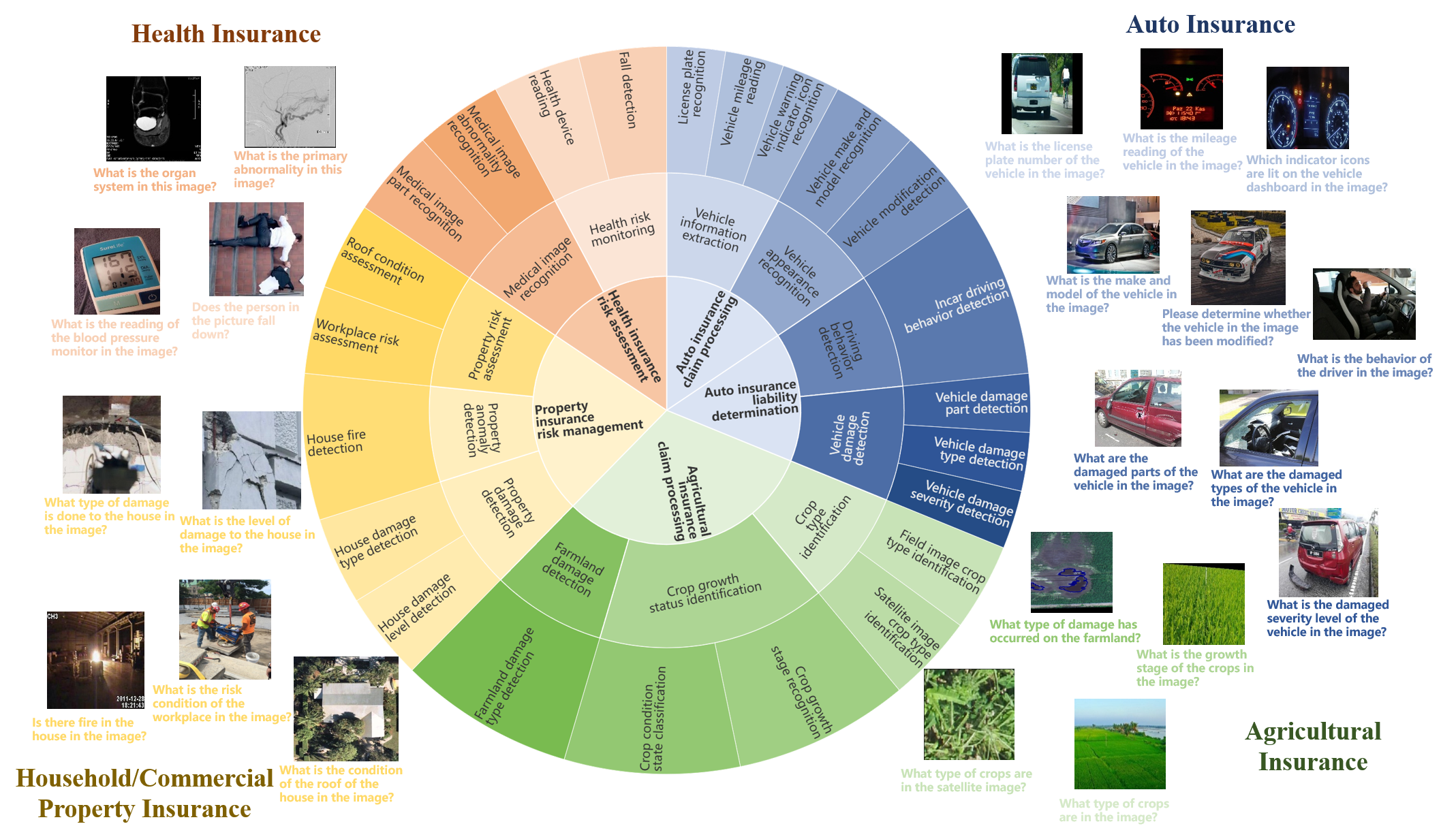}
    \end{subfigure}
    \caption{\textbf{(Left)} The task pyramid of INS-MMBench. \textbf{(Right)} Overview of INS-MMBench. INS-MMBench constructs 5 scenario tasks (represented in the inner circle), 12 meta-tasks (represented in the middle circle) and 22 fundamental tasks (represented in the outer circle) across 4 types of insurance, distinguished by 4 primary colors: \textcolor{blue}{blue}, \textcolor{orange}{orange}, \textcolor{yellow}{yellow}, and \textcolor{green}{green}. For each fundamental task, we provide an example of image-question pair.}
    \label{fig:merged}
\end{figure*}

\section{Introduction}\label{sec:intro}

In recent years, Large Language Models (LLMs) have demonstrated remarkably powerful semantic understanding and conversational capabilities~\citep{wei2022emergent,kasneci2023chatgpt,zhao2023survey,shen2023chatgpt,zhang2022automatic}, profoundly impacting human work and life. Building on this foundation, Large Vision-Language Models (LVLMs) and Multimodal Large Language Models (MLLMs) have taken a further step by mapping and aligning visual and textual features, enabling the processing and interaction with multimodal data~\citep{bai2023qwen,zhu2023minigpt,wang2024visionllm,yin2023survey}. Researchers have found that LVLMs exhibit exceptional performance in general tasks such as image recognition, document parsing, and OCR processing~\citep{yang2023dawn,li2023seed1,xu2023lvlm}. Beyond exploring general capabilities, researchers have also begun to apply LVLMs to various specialized domains such as healthcare~\citep{hu2024omnimedvqa,wang2024surgical}, autonomous driving~\citep{dewangan2023talk2bev,li2024automated} and social media content analysis~\citep{lyu2025gpt,zhang2024somelvlm}, highlighting the broad potential of LVLMs in addressing domain-specific challenges and tasks.

Insurance, as a domain characterized by a wide range of multimodal application scenarios, involves extensive use of multimodal data and computer vision algorithms in practical operations~\citep{fernando2022automated,sahni2020insurance,zhang2020automatic,li2018anti}. This offers substantial opportunities for integrating LVLMs into the insurance industry. For instance, in auto insurance, analyzing images of damaged vehicles can support rapid assessment and accurate damage estimation~\citep{mallios2023vehicle}, while in property insurance, analyzing images of buildings can help evaluate potential risks~\citep{xu2021computer}. However, existing research~\citep{lin2024harnessing} has only qualitatively analyzed LVLM applications in insurance, without systematically organizing relevant multimodal tasks or constructing domain-specific benchmarks. This has hindered the comprehensive assessment and broader adoption of LVLMs in the insurance sector.

To address this challenge, we introduce \textbf{INS-MMBench}, the first comprehensive and hierarchical benchmark designed to evaluate LVLMs in the insurance domain (Figure~\ref{fig:merged}). We begin by systematically organizing and refining a multimodal task framework spanning four representative types of insurance: auto, property, health, and agricultural. This framework is developed using a bottom-up hierarchical methodology and is structured into three levels: \textbf{fundamental task}, which focuses on the understanding of individual insurance-related visual elements; \textbf{meta-task}, which involves the compositional understanding of multiple insurance-related visual elements; and \textbf{scenario task}, which pertains to real-world insurance tasks requiring multi-step reasoning and decision-making. Next, based on the task framework, we execute a benchmark construction pipeline, including data search, processing, and question/answer construction. INS-MMBench includes a total of 12,052 images, 10,372 thoroughly designed questions (including multiple-choice visual questions and free-text visual questions), comprehensively covering 22 fundamental tasks, 12 meta-tasks, and 5 scenario tasks,  aligned with key stages of the insurance process, including underwriting, risk monitoring, and claims processing.

Furthermore, we select 11 LVLMs for evaluation and conduct a comprehensive analysis of the results. The key findings from the evaluation are as follows: (1) None of the evaluated LVLMs achieve a score above 70 on the fundamental tasks. In many cases, their performance \textbf{falls short of the human baseline}, underscoring the inherent complexity and domain-specific challenges of insurance-related multimodal tasks; (2) LVLMs face significant challenges in complex insurance scenarios, particularly in \textbf{multi-step reasoning}, suggesting that future improvements should focus on domain-specific multimodal reasoning datasets and reinforcement learning to enhance reasoning capabilities; (3) There are significant differences in LVLMs' performance across \textbf{different insurance types and tasks}, with better results in auto insurance and health insurance compared to property insurance and agricultural insurance, suggesting that real-world adoption may follow a staged approach, prioritizing areas where LVLMs show the most promise; (4) \textbf{The gap between open-source and closed-source LVLMs is shrinking}, with some open-source models now approaching or even surpassing the capabilities of closed-source models in some tasks such as medical image recognition, vehicle appearance recognition and farmland damage detection, indicating that fine-tuned open-source models may offer a cost-effective and data-secure alternative for insurance applications; and (5) The primary sources of LVLMs' errors are the \textbf{lack of knowledge and reasoning skills in the insurance field}. While prompt engineering offers partial improvement, significant advancements will require targeted optimization and continued research into domain-adapted LVLMs.

In summary, our main contributions are as follows: (1) We conduct a systematic review and categorization of multimodal tasks across key insurance types, using a bottom-up hierarchical task definition methodology; (2) We introduce INS-MMBench, the first hierarchical benchmark designed to evaluate LVLMs in the insurance domain; and (3) We benchmark multiple representative LVLMs, offering insights to inform and guide future research and development of LVLMs for insurance applications.

\section{Related work}

\subsection{Large Vision-Language Models}

With the rapid development of Large Language Models (LLMs)~\citep{chang2024survey, wei2022emergent, huang2022large}, researchers are leveraging the powerful generalization capabilities of these pre-trained LLMs for processing and understanding multimodal data~\citep{ye2023mplug, zhao2023learning, deshmukh2023pengi}. A key area of focus is the use of Large Vision-Language Models (LVLMs) for visual inputs. LVLMs employ visual encoders and visual-to-language adapters to encode the visual features from image data and align these features with textual features. The combined features are then processed by pre-trained LLMs, leading to significant advancements in visual recognition and understanding~\citep{yin2023survey, wu2023multimodal}.

Various open-source and closed-source LVLMs are continuously emerging. In the realm of open-source models, notable examples include LLaMA-Adapter~\citep{zhang2023llama}, LLaVA~\citep{liu2024visual}, BLIP-2~\citep{li2023blip}, MiniGPT-4~\citep{zhu2023minigpt}, and InternVL~\citep{chen2023internvl}. These models have successfully integrated visual and textual modalities, achieving commendable results. In the closed-source domain, representative models include GPT-4o~\citep{GPT4o}, GPT-4V~\citep{achiam2023gpt}, Gemini~\citep{gemini}, and Qwen-VL~\citep{QwenVL}, all of which have demonstrated outstanding performance in numerous tests and evaluations~\citep{yang2023dawn, fu2023challenger, li2023comprehensive}. We evaluate both open-source and closed-source LVLMs to verify the capability of different models in the insurance domain.

\subsection{Benchmarks for LVLMs}

\begin{table}[t]
    \centering
    \caption{Comparison of Different Benchmark Datasets.}
    \label{benchmark_comparison}
    \adjustbox{max width=0.48\textwidth}{
    \begin{tabular}{lcccc}
        \toprule
        \textbf{Dataset} & \textbf{Type} & \textbf{Size} & \textbf{Models} & \textbf{Potential Overlap} \\
        \midrule
        INS-MMBench (Ours)       & Domain-specific: insurance & 12,052  & 11 & - \\
        SEED-Bench~\citep{li2024seed}       & Comprehensive              & 19,242   & 18 & No \\
        MMBench~\citep{liu2023mmbench}          & Comprehensive              & 2,974    & 14 & No \\
        SciFIBench~\citep{roberts2024scifibench}       & Task-specific: scientific images & 1,000  & 29 & No \\
        MMC-Benchmark~\citep{liu2023mmc}    & Task-specific: charts      & 2,000    & 6  & No \\
        OmniMedVQA~\citep{hu2024omnimedvqa}       & Domain-specific: medical      & 127,995 & 12 & Yes  \\
        Mathvista~\citep{lu2023mathvista}        & Domain-specific: math   & 5,487    & 9  & No \\
        \bottomrule
    \end{tabular}}
\end{table}

As research into LVLMs accelerates, an increasing number of researchers are proposing benchmarks to evaluate the capabilities of  models~\citep{ye2023mplug, zhang2024m3exam, liu2023mitigating, chen2024we}. Based on the scope of capability evaluation, these studies can be categorized into three types: task-specific, comprehensive, and domain-specific benchmarks.

\textbf{Comprehensive benchmarks} aim to evaluate LVLMs across a broad spectrum of general capabilities and tasks. These benchmarks are typically constructed by systematically defining and categorizing the core competencies of LVLMs, enabling wide-ranging and holistic assessments. Representative studies include LVLM-eHub~\citep{xu2023lvlm}, SEED-Bench~\citep{li2023seed1, li2023seed2}, MMBench~\citep{liu2023mmbench}, MME, and MMT-Bench~\citep{ying2024mmt}.

\textbf{Task-specific benchmarks} focus on particular tasks and types of visual data, providing detailed task definitions. Examples include SciFIBench~\citep{roberts2024scifibench} for scientific images, MMC-Benchmark~\citep{liu2023mmc} for charts, MVBench~\citep{li2023mvbench} (using video frames as input) for videos and SEED-Bench-2-Plus~\citep{li2024seed} for web pages, charts and maps.

\textbf{Domain-specific benchmarks} are designed for visual tasks within specific professional domains. Given the unique knowledge requirements and task structures in these fields, general-purpose benchmarks often fall short in effectively evaluating LVLM performance. To address this, researchers have developed specialized benchmarks for domains such as healthcare (OmniMedVQA~\citep{hu2024omnimedvqa}), mathematics~\citep{lu2023mathvista, wang2024measuring}, autonomous driving (Talk2BEV-Bench~\citep{dewangan2023talk2bev}), and geography~\citep{roberts2023charting}. However, as previously noted, the insurance domain and even the broader finance domain still lack corresponding domain-specific benchmarks for LVLM evaluation~\citep{chen2024fintextqa,li2023large,lin2024harnessing}. To fill this gap, we introduce INS-MMBench, a domain-specific benchmark designed to advance the development and application of LVLMs in the insurance industry.

As shown in Table~\ref{benchmark_comparison}, we conduct a detailed comparison based on the three benchmark categories outlined above. Six relevant benchmarks are examined across  dimensions including benchmark type, dataset size, number of evaluated models, and potential overlap with our benchmark.

\section{INS-MMBench}\label{sec:INS-MMBench}

\subsection{Task framework}

Given the differences in practical workflows among various types of insurance, we select four core types—auto, property, health, and agricultural insurance—for our benchmark. These categories cover both life and property insurance, which dominate the market and are widely representative~\citep{weedige2019decision,driver2018insurance}. Moreover, they feature distinct multimodal tasks closely tied to real-world applications. For instance, auto insurance involves the assessment of vehicle damage through visual inspection, while property insurance covers evaluations of damaged buildings or personal property.

To ensure that our benchmark closely aligns with real-world insurance applications and is capable of evaluating the capabilities of LVLMs in this context, we have developed a bottom-up hierarchical task definition methodology. Based on this methodology, we have constructed a systematic multimodal task framework tailored for the insurance sector. This framework incorporates a three-layer hierarchy:
\textbf{Fundamental task} focuses on recognizing and understanding an individual insurance-related visual element.
\textbf{Meta-task} combines multiple related fundamental tasks to enable integrated understanding and reasoning of interconnected visual elements.
\textbf{Scenario task} simulates real-world insurance application that requires multi-step reasoning and decision-making, thereby capturing the complexity and dynamics inherent in actual operations.

\begin{figure*}[!htb]
    \centering
    \includegraphics[width=0.7\linewidth]{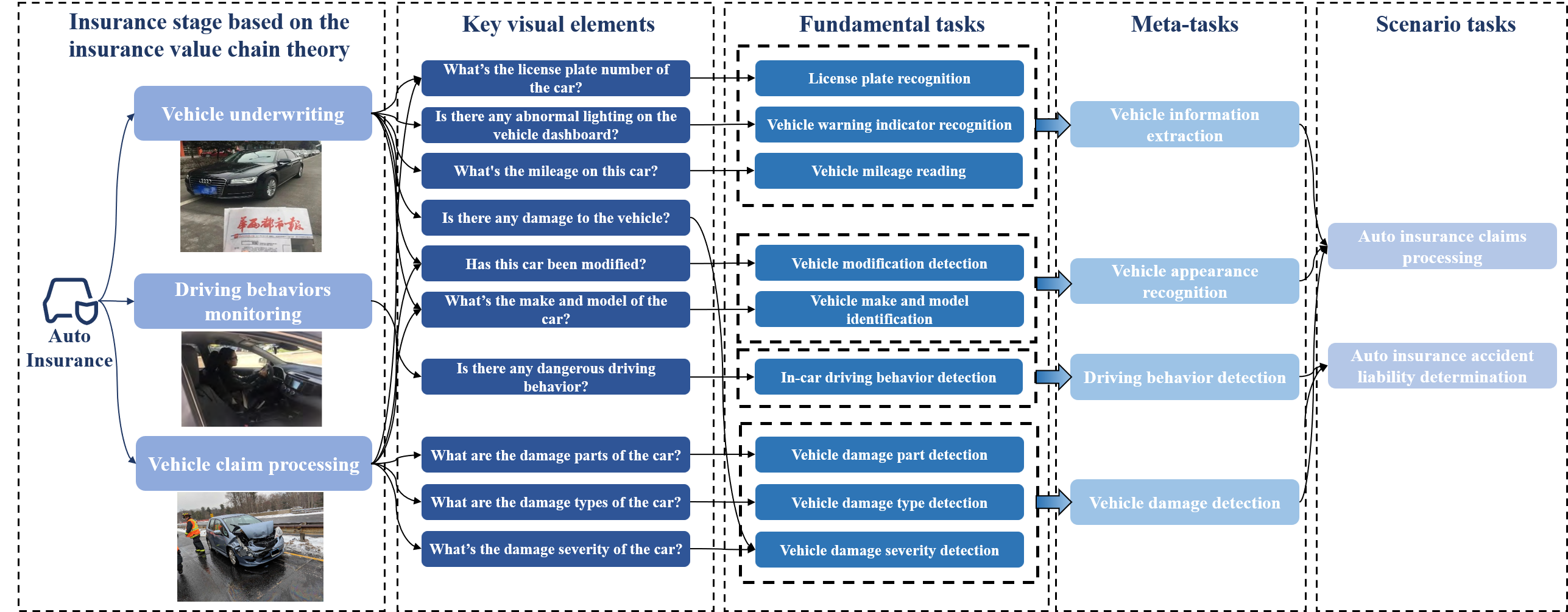}
    \caption{An illustration of our bottom-up hierarchical task definition methodology. First, we identify and categorize different insurance stages. Next, we enumerate the key visual elements required at each stage. Based on these key visual elements, we define the fundamental tasks. Next, we cluster the fundamental tasks into meta-tasks. Finally, we construct scenario tasks by integrating multiple meta-tasks into realistic, end-to-end insurance situations that require multi-step reasoning.}
    \label{task_construction_process}
\end{figure*}

\begin{figure*}
    \centering
    \includegraphics[width=0.7\linewidth]{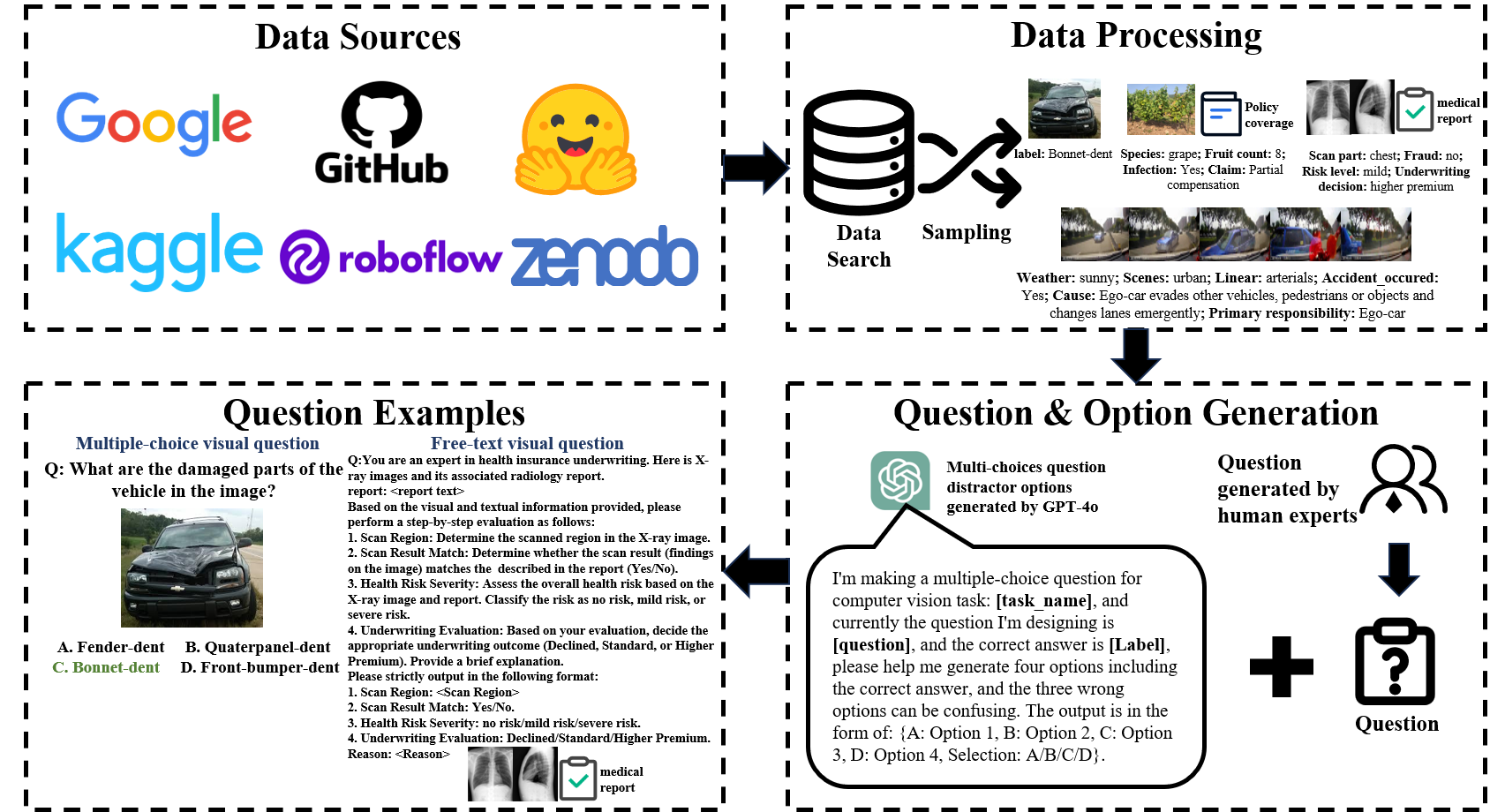}
    \caption{An illustration of our data collection and benchmark construction process. First, we collect datasets from multiple public sources. Next, we perform manual filtering and sampling of the datasets, followed by the necessary data processing. Finally, human experts construct both multiple-choice and free-text visual questions, while GPT-4o is used to generate answer options for a subset of the multi-choice questions.}
    \label{pipeline}
\end{figure*}

As an example, we discuss the detailed task construction process for auto insurance (Figure~\ref{task_construction_process}). 
Initially, based on the \textbf{insurance value chain theory}~\citep{eling2018impact, eling2022impact}, we select three key stages that are rich in multimodal data and tasks: vehicle underwriting, vehicle risk monitoring, and vehicle claim processing. At each stage, we identify the key visual elements that insurance operators need to extract. For instance, during the vehicle underwriting stage, operators must identify license plate information, vehicle model, dashboard readings, and vehicle condition. These are essential for information collection, condition verification, and underwriting decision. Based on these key visual elements, we define the fundamental tasks. For example, the requirement to recognize license plate information motivates the definition of the license plate recognition task. By following this process, we define a total of 9 fundamental tasks for auto insurance. Next, we cluster these fundamental tasks based on their characteristics to form 4 meta-tasks. Finally, drawing on real-world insurance scenarios, we integrate various meta-tasks to construct 2 end-to-end insurance scenario tasks. Following this approach, we have developed a framework consisting of 22 fundamental tasks, 12 meta-tasks, and 5 scenario tasks across the 4 insurance categories.

\subsection{Benchmark construction}

The details of our data collection and benchmark construction process (Figure~\ref{pipeline}) are as follows:

\textbf{Data sources.} We search for datasets relevant to each task in our framework by issuing task‑specific keywords across several popular public sources, including Google, Kaggle, GitHub, Hugging Face, Zenodo, and Roboflow. When multiple datasets are available for a task, we compare candidates using usage statistics and community feedback, and retain those that best match our needs. Guided by human experts, we prioritize datasets with high applicability to real insurance scenarios, as summarized in Table~\ref{insurance_table}.

\begin{table*}
  \caption{Overview of the datasets used in INS-MMBench. The \textit{Images} and \textit{Questions} columns list, respectively, the number of images and the number of questions for each task. The \textit{Type} column denotes the question format, where \textbf{choice} refers to multiple-choice visual questions, and \textbf{text} refers to free-text visual questions.}
  \label{insurance_table}
  \centering
  \adjustbox{max width=0.85\textwidth}{
  \begin{tabular}{clllccc}
    \cline{1-7}
    \textbf{Insurance type} & \textbf{Meta-tasks / Scenario tasks name} & \textbf{Fundamental tasks name} & \textbf{Dataset} & \textbf{Images} & \textbf{Questions} &\textbf{Type}\\
    \cline{1-7}
    \multirow{1}{*}{Auto insurance} & Vehicle information extraction & License plate recognition & \begin{tabular}[c]{@{}l@{}} CCPD~\citep{xu2018towards}, \\ mjdfodf-qmbuf~\citep{mjdfodf-qmbuf_dataset}\end{tabular} & 250 & 250 & choice\\
    & & Vehicle mileage reading & TRODO~\citep{mouheb2021trodo} & 500 & 500 & choice\\
    & & Vehicle warning indicator recognition & dataset\_dashboard~\citep{dataset_dashboard__dataset} & 500 & 500 & choice\\
    \cline{2-7}
    & Vehicle appearance recognition & Vehicle make and model identification & Stanford Cars~\citep{krause20133d} & 500 & 500 & choice\\
    & & Vehicle modification detection & tuning-car-detection~\citep{tuning-car-detection_dataset} & 100 & 100 & choice\\
    \cline{2-7}
    & Driving behavior detection & Incar driving behavior detection & \begin{tabular}[c]{@{}l@{}} Driver-Distraction-Dataset~\citep{ezzouhri2021robust}\end{tabular} & 500 & 500 & choice \\
    \cline{2-7}
    & Vehicle damage detection & Vehicle damage part detection & car\_dent\_scratch\_detection-1~\citep{car_dent_scratch_detection-1_dataset} & 500 & 500 & choice\\
    & & Vehicle damage type detection & Cardd~\citep{wang2023cardd} & 500 & 500 & choice\\
    & & Vehicle damage severity detection & \begin{tabular}[c]{@{}l@{}} car-crash-severity-detection\citep{car-crash-severity-detection_dataset}\end{tabular} & 308 & 308 & choice\\
    \cline{2-7}
    & Auto insurance claim processing & \texttimes  & DADA~\citep{fang2021dada} & 1350 & 270 & text\\
    \cline{2-7}
    & Auto insurance accident liability determination & \texttimes  & Fast Furious and Insured~\citep{HackerEarthInsurance} & 400 & 400 & text \\
    \cline{1-7}
    \multirow{1}{*}{Property insurance} & Property risk assessment & Roof condition assessment & damages-svll3~\citep{damages-svll3_dataset} & 500 & 500 & choice\\
    & & Workplace risk assessment & worker-safety~\citep{worker-safety_dataset} & 100 & 100 & choice\\
    \cline{2-7}
    & Property anomaly detection & House fire detection & fire-detection-cta61~\citep{fire-detection-cta61_dataset} & 498 & 498 & choice\\
    \cline{2-7}
    & Property damage detection & House damage type detection & damage-type~\citep{damage-type_dataset} & 469 & 469 & choice\\
    & & House damage level detection & damage-level~\citep{damage-level_dataset} & 409 & 409 & choice\\
    \cline{2-7}
    & Property insurance risk management & \texttimes  & xBD~\citep{gupta2019creating} & 400 & 200 & text\\
    \cline{1-7}
    \multirow{1}{*}{Health insurance} & Health risk monitoring & Fall detection & \begin{tabular}[c]{@{}l@{}} Fall Detection Dataset\citep{fall_detection_detection}\end{tabular} & 374 & 374 & choice\\
    & & Health device reading & blood-pressure-monitor-display~\citep{blood-pressure-monitor-display_dataset} & 100 & 100 & choice\\
    \cline{2-7}
    & Medical image recognition & Medical image organ recognition & VQA-Med 2019~\citep{abacha2019vqa} & 500 & 500 & choice\\
    & & Medical image abnormality recognition & VQA-Med 2019~\citep{abacha2019vqa} & 500 & 500 & choice\\
    \cline{2-7}
    & Health insurance risk assessment & \texttimes  & IU-XRay~\citep{demner2016preparing} & 800 & 400 & text\\
    \cline{1-7}
    \multirow{1}{*}{Agricultural insurance} & Crop type identification & Field image crop type identification & \begin{tabular}[c]{@{}l@{}} agricultural crop images\citep{Agriculturecropimages} \end{tabular} & 250 &250 & choice\\
    & & Satellite image crop type identification & \begin{tabular}[c]{@{}l@{}}Drone Imagery Classification\\ Training Dataset for Crop Types\\ in Rwanda~\citep{chew2020deep}\end{tabular} & 498 & 498 & choice\\
    \cline{2-7}
    & Crop growth status identification & crop growth stage recognition & wheat-growth-stage-challenge~\citep{WheatGrowthStageChallenge} & 500 & 500 & choice\\
    \cline{2-7}
    & Farmland damage detection & Farmland damage type detection & agriculture-vision~\citep{chiu2020agriculture} & 500 & 500 & choice\\
    \cline{2-7}
    & Agricultural insurance claim processing & \texttimes  & Precision viticulture dataset~\citep{velez_2023_10362568} & 246 & 246 & text \\
    \cline{1-7}
     Total & & & & 12052 & 10372 & \\
    \cline{1-7}
  \end{tabular}}
\end{table*}

\textbf{Data processing.} This stage comprises two key subtasks: sampling and structuring.

\begin{itemize}
  \item \textbf{Sampling}: We apply a carefully designed sampling strategy. For classification tasks with few labels (fewer than 10 in our case), we use stratified sampling to ensure balanced representation and minimize bias. For tasks with more diverse labels, we adopt random sampling to capture a broad spectrum of responses. To balance coverage with the cost of LVLMs evaluation, we sample up to 500 instances per task, or retain the entire dataset when it contains fewer than 500 instances. Details of the selected datasets are provided in Table~\ref{insurance_table}.
  \item \textbf{Structuring}: The raw inputs fall into three formats: (1) an image with a single label, (2) multiple images with multiple labels, and (3) multiple images accompanied by additional textual information (\textit{e.g.,} policies or medical reports) with multiple labels. We convert each dataset into a unified CSV file that records the image path(s), any accompanying text, and the corresponding label(s) for downstream use.
\end{itemize}

\textbf{Question and answer construction.} We adopt two question formats: multiple-choice visual questions for fundamental tasks and free-text visual questions for scenario tasks. All questions are crafted under the guidance of insurance experts. The construction of answers, however, depends on the question format. For multiple-choice visual questions, if the underlying dataset has four or fewer labels, those labels are used as the options (\textit{e.g.,} the four levels of Vehicle Damage Severity: no accident, minor damage, moderate damage, and severe damage). When datasets have more complex or open‑ended labels, we retain the correct label and employ GPT‑4o to generate semantically plausible distractors, thereby completing the option set. For free-text visual questions, we decompose complex insurance scenario into a structured, step-by-step reasoning process and manually construct reference answers under the guidance of experts to reflect real-world decision-making.
Following this process, INS-MMBench comprises 12,052 images and 10,372 questions.

\section{Experiment}\label{sec:experiment}

\subsection{Experimental setting}
Our evaluation proceeds in two stages. In the first stage, we test every selected LVLM on the fundamental tasks in INS‑MMBench, obtaining a broad snapshot of each model’s basic multimodal understanding and reasoning abilities in insurance domain. In the second stage, we take the top‑performing LVLMs from stage one and subject them to scenario tasks that emulate complete, end‑to‑end insurance workflows. These tasks involve multi-step visual–text reasoning and offer insight into the models’ potential performance in real-world applications.

\textbf{Selected LVLMs.} We select a representative set of 11 LVLMs for our evaluation. This set includes 8 closed-source LVLMs: GPT-4o, GPT-4V, GPT-4o-mini, Gemini 1.5 Flash, Qwen-VL-Plus, Qwen-VL-Max, GLM4V-Plus-0111 and Claude3V-Haiku as well as 3 open-source LVLMs including LLaVA, Qwen-2.5-VL-32B and Qwen-VL-Chat.

\begin{table*}
    \centering
    \caption{Evaluation results of the LVLMs across different meta-tasks. The values in the table represent the average accuracy. Specifically, \textbf{VIE} denotes vehicle information extraction, \textbf{VAR} denotes vehicle appearance recognition, \textbf{DBD} denotes driving behavior detection, \textbf{VDD} denotes vehicle damage detection, \textbf{HPAD} denotes household/commercial property anomaly detection, \textbf{HPDD} denotes household/commercial property damage detection, \textbf{HPRA} denotes household/commercial property risk assessment, \textbf{HRM} denotes health risk monitoring, \textbf{MIR} denotes medical image recognition, \textbf{CGSI} denotes crop growth stage identification, \textbf{CTI} denotes crop type identification, \textbf{FDD} denotes farmland damage detection. The highest and second-highest results are highlighted in \textbf{bold} and \underline{underlined}, respectively.}
    \label{evaluation_results_detail}
    \adjustbox{max width=0.75\textwidth}{
    \begin{tabular}{lcccccccccccc}
    \toprule
    \textbf{Model}  & \multicolumn{1}{l}{\textbf{VIE}} & \multicolumn{1}{l}{\textbf{VAR}} & \multicolumn{1}{l}{\textbf{DBD}} & \multicolumn{1}{l}{\textbf{VDD}} & \multicolumn{1}{l}{\textbf{HPAD}} & \multicolumn{1}{l}{\textbf{HPDD}} & \multicolumn{1}{l}{\textbf{HPRA}} & \multicolumn{1}{l}{\textbf{HRM}} & \multicolumn{1}{l}{\textbf{MIR}} & \multicolumn{1}{l}{\textbf{CGSI}} & \multicolumn{1}{l}{\textbf{CTI}} & \multicolumn{1}{l}{\textbf{FDD}} \\
    \midrule
    GPT-4o           & { \textbf{81.12}}             & { \textbf{98.50}}             & {\ul {88.60}}             & {\ul {83.94}}                            &  \textbf{91.16}              & {\textbf{47.04}}              & 65.50              & {\textbf{95.72}}             & {\ul{66.50}}             & 30.80              & {\textbf{41.31}}             & {\ul {34.60}}             \\
    Qwen-VL-Max        & 75.28                            & {\ul {98.20}}                            & 74.80                            & 81.88             & 80.72                             & 45.79                             & {\textbf{71.80}}                             & 88.24                            & 64.00                            & 29.60                             & {\ul {40.37}}                   & 26.00                            \\
    Gemini 1.5 Flash  & 67.28                            & 96.80             & 79.20                            & \textbf{84.40}                            & 74.30              & 46.36                             & { \ul {70.40}}                             & 81.82                            & 66.00                            & {\ul {36.60}}                             & 38.10                            & 21.20                            \\
    GLM4V-Plus-0111 & 77.68 & 92.60 & 84.40 & 80.58 & 58.84 & 42.94 & 67.00 & 90.11 & 65.90 & 28.80 & 33.56 & 25.60 \\
    
    GPT-4V & 72.16                            & 93.60                            & 66.20                            & 80.35                            & 88.35                             & 41.80                             & 65.80                             & 94.12                            & 62.10                            & 23.60                             & 39.17                            & 20.00                   \\
    
    GPT-4o-mini        & 70.24                           & 95.20                            & 85.80                            & 75.23             & {\ul {89.56}}                             & 39.75                             & 60.60                             & {\ul {94.39}}                            & 52.10                            & 23.80                            & 34.36                   & 15.00                            \\
    Qwen-VL-Plus       & 63.84                            & 96.20                            & 69.60                            & 69.88                            & 57.03                             & 39.18                             & 56.40                             & 86.10                            & 57.00                            & 15.40                             & 25.40                            & 18.20                            \\
    
    Claude3V-Haiku & 45.76                             & 86.8                             & 52.40                            & 66.13                             & 75.10                             & 27.90                             & 62.40                             & 84.49                            & 49.50                            & 19.80                             & 23.53                            & 7.60                   \\
    \midrule
    Qwen-2.5-VL-32B & 76.48 & 94.80 & 75.00 & 77.83 & 73.69 & 38.04 & 64.60 & 91.98 & \textbf{70.40} & 27.60 & 38.90 & 32.00 \\
    Qwen-VL-Chat & 44.32                             & 94.60                             & 59.60                             & 55.50                             & 59.04                              & 30.41                              & 60.00                              & 80.75                             & 59.30                             & 15.80                              & 30.62                             & 13.00                    \\
    LLaVA &  32.64  &  60.20  &  51.80  &  49.69  &  87.35  &  34.85  &  65.00  &  83.69  &  57.54  &  21.40  &  37.57  &  14.20 \\
    \midrule
    Human baseline & {\ul {79.50}}                            & 59.50                            & \textbf{98.67}                            &  52.78                            & 73.33                             & {\ul {46.67}}                             & 63.33                             & 85.00                            & 65.00                            & \textbf{60.00}                             & 35.00                            & \textbf{40.00}                   \\
    \bottomrule
    \end{tabular}}
\end{table*}

\begin{table}
    \centering
    \caption{Evaluation results of the LVLMs across different insurance types. The values in the table represent the average accuracy. The highest and second-highest results are highlighted in \textbf{bold} and \underline{underlined}, respectively.}
    \label{evaluation_results}
    \adjustbox{max width=0.47\textwidth}{
    \begin{tabular}{lccccc}
        \toprule
        \textbf{Model}  & \textbf{Overall}     & \textbf{\begin{tabular}[c]{@{}c@{}}Auto\\      insurance\end{tabular}} & \textbf{\begin{tabular}[c]{@{}c@{}}Household/commercial\\      property insurance\end{tabular}} & \textbf{\begin{tabular}[c]{@{}c@{}}Health\\      insurance\end{tabular}} & \textbf{\begin{tabular}[c]{@{}c@{}}Agricultural\\      insurance\end{tabular}} \\
        \midrule
        GPT-4o          & {\textbf{69.70}} & {\textbf{86.00}}      & {\textbf{63.77}}                             & {\textbf{76.73}}        & {\ul{36.38}}           \\
        Qwen-VL-Max       & {\ul{65.33}}                & 80.86                     & {\ul{61.99}}                                            & 70.60                       & 33.18                           \\
        Gemini 1.5 Flash &  64.21                & {\ul{79.40}}                     & 60.18                                            & 70.31                       & 32.84 \\
        GLM4V-Plus-0111 & 63.51 & {\ul{81.79}} & 53.57 & 72.49 & 29.92 \\
        GPT-4V      & 62.79                & 77.35                     & 60.55                                           & 70.82                       & 29.23       \\

        GPT-4o-mini       & 60.66                & 77.77                     & 58.53                                            & 63.61                       & 25.80                           \\
        Qwen-VL-Plus      & 54.94                & 71.42                     & 48.51                                            & 64.92                       & 20.48       \\

        Claude3V-Haiku      & 48.95                & 59.95                     & 49.63                                             & 59.02                        & 17.91        \\
        \cline{1-6}
        
        Qwen-2.5-VL-32B & 64.10 & 79.34 & 54.58 & {\ul{76.27}} & 33.70 \\
        Qwen-VL-Chat      & 48.85                & 57.64                     & 45.90                                            & 65.14                       & 21.34       \\
        LLaVA &  46.99  &  45.47  &  56.82  &  65.25  &   26.26\\
        \midrule 
        Human baseline      & 60.45               & 62.22                    & 60.00                                            & 75.00                       & \textbf{42.50}       \\ 
        \bottomrule
    \end{tabular}}
\end{table}

\textbf{Evaluation methods.} We employ VLMEvalKit, an open-source evaluation toolkit for LVLMs~\citep{duan2024vlmevalkit}, to conduct our evaluation. This toolkit supports integrated testing of both closed-source and open-source LVLMs and is adaptable to custom benchmarks. We extend VLMEvalKit to suit the two question formats in INS‑MMBench. For multiple-choice visual questions, we utilize the combination methods provided by VLMEvalKit—exact matching (identifying ``A," ``B," ``C," or ``D" in the output strings) and LLM-based answer extraction (analyzing responses using GPT-4o). For multi-step reasoning free-text visual questions, we first parse each response into explicit reasoning steps and a final decision, then verify every part: classification steps are graded by regex‑based exact matching, while numerical steps (such as predicted insurance premiums) are scored by the mean deviation from ground‑truth values. This step‑wise grading scheme enables us to measure both the reasoning process and the final outcome, yielding a more nuanced picture of each LVLM’s performance in the insurance domain.

\textbf{Human baseline experiment.} Additionally, we conduct a human baseline experiment involving three graduate students specializing in insurance and three industry professionals from leading insurance companies. The industry professionals come from diverse departments, including actuarial, business operations, and strategic development. Together, the participants are tasked with answering a subset of 220 questions (10 from each fundamental task) selected from our benchmark. This approach leverages both academic expertise and practical industry insights, providing a well-rounded evaluation.

\subsection{Main results}

Tables~\ref{evaluation_results_detail} and \ref{evaluation_results} present the stage-one evaluation results of LVLMs across meta-tasks and insurance types, respectively. The results are organized into three sections: the first eight rows present the evaluation results of closed-source models, the middle three rows show the evaluation results of open-source models, and the last row provides the human baseline. We then select the top-performing models—GPT-4o, Qwen-VL-Max, and Gemini 1.5 Flash—for further evaluation on five scenario tasks. The stage-two results are presented in Table~\ref{tab:insurance_tasks}. Based on the results shown in Tables~\ref{evaluation_results_detail}, \ref{evaluation_results} and \ref{tab:insurance_tasks}, we draw the following observations.

\begin{table}
    \centering
    \caption{Evaluation results of selected LVLMs across different scenario tasks. The evaluation metrics include two types: (1) Accuracy (\%) – metrics evaluating correctness in percentage format, and (2) Difference (Diff) – numerical metrics representing the average deviation between true and predicted values. Rows with a \colorbox{yellow!20}{yellow} background indicate the final insurance decision for each scenario.}
    \label{tab:insurance_tasks}
    \resizebox{0.48\textwidth}{!}{ 
    \begin{tabular}{p{4.5cm} l c c c} 
        \toprule
        \textbf{Scenario Tasks} & \textbf{Reason Step} & \textbf{GPT-4o} & \textbf{Qwen-VL-Max} & \textbf{Gemini 1.5 Flash} \\
        \midrule
        \multirow{5}{=}{\raggedright Auto Insurance Claim Processing} 
        & Damage judgment (\%) & 82.75 & 86.75 & 82.00 \\
        & Damage Severity (\%) & 30.50 & 29.00 & 25.50 \\
        & Estimated Loss (Diff) & 6185.61 & 9584.37 & 10894.78 \\
        & Claim Eligibility (\%) & 62.00 & 55.50 & 47.50 \\
         & \cellcolor{yellow!20} Final Claim Decision (Diff) & \cellcolor{yellow!20}3029.09 & \cellcolor{yellow!20}3686.32 & \cellcolor{yellow!20}4258.76 \\
        \midrule
        \multirow{6}{=}{\raggedright Auto Insurance Accident Liability Determination} 
        & Weather Classification (\%) & 71.11 & 64.07 & 71.11 \\
        & Scene Classification (\%) & 81.85 & 81.85 & 84.07 \\
        & Linear Classification (\%) & 61.11 & 54.44 & 56.30 \\
        & Accident Occurrence (\%) & 55.19 & 31.48 & 14.81 \\
        & Accident Cause judgment (\%) & 24.80 & 9.63 & 3.33 \\
         & \cellcolor{yellow!20}Responsible Party Identification (\%) & \cellcolor{yellow!20}9.26 & \cellcolor{yellow!20}5.56 & \cellcolor{yellow!20}2.59 \\ 
        \midrule
        \multirow{4}{=}{\raggedright Health Insurance Risk Assessment} 
        & Scan Region Classification (\%) & 94.00 & 94.25 & 100.00 \\
        & Fraud detection (\%) & 50.75 & 56.75 & 57.75 \\
        & Health Risk Assessment (\%) & 84.07 & 83.19 & 85.80 \\
         & \cellcolor{yellow!20}Underwriting Decision (\%) & \cellcolor{yellow!20}43.50 &\cellcolor{yellow!20} 22.00 &\cellcolor{yellow!20} 48.25 \\
        \midrule
        \multirow{5}{=}{\raggedright Property Insurance Risk Management} 
        & Disaster Occurrence judgment (\%) & 67.00 & 71.50 & 69.00 \\
        & Disaster Type Classification (\%) & 34.00 & 32.50 & 37.00 \\
        & Properties Count (Diff) & 54.64 & 43.10 & 38.77 \\
        & Damaged Properties Count (Diff) & 18.79 & 21.03 & 17.86 \\
        &  \cellcolor{yellow!20}Insurance Decision (\%) &\cellcolor{yellow!20} 51.00 & \cellcolor{yellow!20}44.50 & \cellcolor{yellow!20}41.00 \\
        \midrule
        \multirow{4}{=}{\raggedright Agricultural Insurance Claim Processing} 
        & Crop Species Classification (\%) & 100.00 & 100.00 & 97.97 \\
        & Fruit Count (Diff) & 7.76 & 6.83 & 11.57 \\
        & Pest Infection judgment (\%) & 87.80 & 85.37 & 72.36 \\
         &  \cellcolor{yellow!20}Claim Decision (\%) & \cellcolor{yellow!20}84.15 &\cellcolor{yellow!20} 81.30 & \cellcolor{yellow!20}68.70 \\
        \bottomrule
    \end{tabular}}
\end{table}

\textbf{GPT-4o leads in performance but highlights the challenges of insurance tasks for LVLMs.} Overall, GPT-4o outperforms all other models, achieving the highest score of 69.70. However, most LVLMs still fall short of or only marginally exceed the human baseline across many tasks, highlighting the complexity and difficulty of insurance-specific scenarios. These findings suggest substantial room for improvement in adapting LVLMs for real-world applications in the insurance domain.

\textbf{LVLMs show substantial performance variance across  meta-tasks and insurance types.} Experimental results show that both open-source and proprietary LVLMs perform better on tasks related to auto and health insurance (\textit{e.g.,} GPT-4o scores 86.00 and 76.73, respectively) than on those involving property and agricultural insurance (\textit{e.g.,} GPT-4o scores drop to 63.77 and 36.38). A more in-depth analysis indicates that this variance stems not only from differences across insurance types but also from the inherent characteristics of the meta-tasks. LVLMs excel in tasks that rely heavily on visual perception and object detection (\textit{e.g.,} vehicle information extraction, vehicle appearance recognition). In contrast, their performance declines significantly in more complex tasks that demand domain-specific knowledge and multi-step reasoning, such as household/commercial property damage detection and crop growth stage identification. These findings suggest that the application of LVLMs in the insurance industry should adopt a phased strategy, initially focusing on tasks where the LVLMs demonstrate strong capabilities, such as auto and health insurance. Meanwhile, further advances in domain adaptation and reasoning enhancement are needed to improve their performance in more specialized tasks and insurance types.

\textbf{LVLMs face significant challenges in complex insurance scenarios.} Although LVLMs demonstrate strong performance in some fundamental tasks and certain reasoning stages of scenario tasks, there remains a substantial gap between their current capabilities and the requirements of real-world scenarios. This is evident from the fact that LVLMs achieve an accuracy exceeding 80\% only on the agricultural insurance claim processing task, while their performance on other scenario tasks falls below 50\%. These findings highlight two key challenges. First, existing LVLMs struggle with multi-step reasoning, suggesting that their near-term application in the insurance industry should be limited to specific fundamental tasks where their performance is more reliable. Second, the analysis of stepwise reasoning provides insights into critical weaknesses of LVLMs. For instance, in the auto insurance accident liability determination task, LVLMs struggle to accurately identify accident occurrences and causes, leading to poor performance in the final insurance decision. Future improvements may come from constructing domain-specific multimodal reasoning datasets for the insurance field and applying reinforcement learning to address reasoning deficiencies.

\textbf{The performance gap between open-source and closed-source LVLMs is narrowing.} Although a noticeable disparity remains, results from fundamental tasks in INS-MMBench show that some open-source LVLMs (\textit{e.g.,} Qwen-2.5-VL-32B) are approaching the performance levels of their proprietary counterparts. This trend suggests that, with continued advancements in model development and increased availability of domain-specific data, high-performing, domain-adapted open-source LVLMs could play a critical role in enabling scalable and cost-effective applications in the insurance industry.

\subsection{Error analysis and mitigation}

To further understand the limitations of LVLMs in the insurance domain, we conduct an in-depth analysis of the errors made by selected models during stage one. Specifically, we examine the error patterns of three LVLMs: GPT-4o, Gemini 1.5 Flash, and Qwen-VL-Max, and categorize the errors into four types: \textbf{perception errors} (where LVLMs do not recognize or detect objects or content within an image), \textbf{lack of insurance knowledge or reasoning ability} (where LVLMs can perceive the visual content but lack the necessary insurance knowledge or reasoning skills to answer the question correctly), \textbf{refusals to answer} (where LVLMs decline to respond to questions they deem sensitive or illegal), and \textbf{lack of adherence to instructions} (where LVLMs do not adhere to the task instructions, resulting in irrelevant responses).

\begin{figure}
    \centering
    \includegraphics[width=1\linewidth]{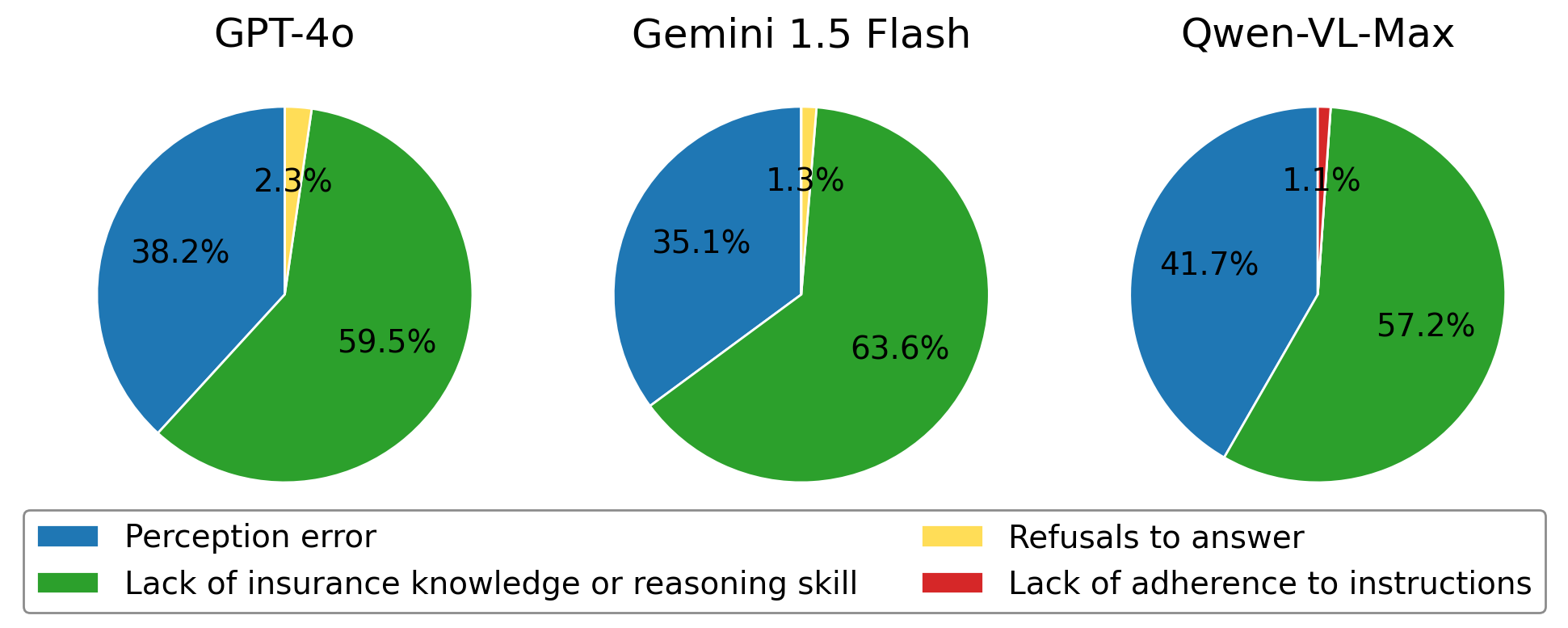}
    \caption{The distribution of error types for GPT-4o, Gemini 1.5 Flash, and Qwen-VL-Max.}
    \label{error_analysis}
\end{figure}

The results of error analysis are illustrated in Figure~\ref{error_analysis}. The most frequent error type is the lack of insurance knowledge or reasoning ability, accounting for 59.5\%, 63.6\%, and 57.2\% of the total errors in GPT-4o, Gemini 1.5 Flash, and Qwen-VL-Max, respectively. Due to insufficient specialized knowledge and analytical skills in the insurance field, LVLMs often struggle to accurately evaluate critical factors such as risk conditions and the extent of damage. Perception errors represent the second most prevalent error type. Constrained by the capabilities of their visual encoders, LVLMs often cannot fully recognize and capture detailed content within images, leading to misclassification or misunderstanding. For instance, GPT-4o misinterprets an image of damaged farmland as ``\textit{an abstract or close-up view of a textured surface with blue and purple hues}." This kind of error is common across all evaluated models.

To address the challenge of limited domain-specific knowledge and reasoning capabilities in the insurance field, we adopt prompt engineering as a mitigation strategy. Specifically, we integrate additional insurance-related information into the original prompts to supplement LVLMs' knowledge and support their analytical reasoning. We evaluate five LVLMs on three fundamental tasks, using 100 randomly sampled instances per task. As shown in Table~\ref{enhanced_prompt_results}, enhanced prompts significantly improve performance in most cases. However, in the vehicle damage detection task for Qwen-VL-Max and Qwen-VL-Plus, the inclusion of conflicting information occasionally confuses the models and reduces accuracy. These findings demonstrate that while prompt engineering is an effective approach, it has inherent limitations, highlighting the need for further enhancement of LVLMs’ domain-specific knowledge and reasoning capabilities to enable more robust insurance applications.

\subsection{Ablation study}
In some of our fundamental tasks, we use GPT-4o to generate distractors for multiple-choice visual questions. This approach may introduce potential bias or insufficient adversarial strength, thereby reducing the confidence in the evaluation results. To address this concern, we conduct additional ablation experiments. Specifically, we select three tasks that require model-generated distractors, comprising a total of 1,500 questions. For these tasks, we generate distractors using Claude 3.5 Haiku and DeepSeek-V3, and evaluate the performance of GPT-4o, Qwen-VL-Max, and Gemini 1.5 Flash. The results (Table~\ref{tab:different_distractor}) indicate that distractors generated by different models can affect the absolute scores of the evaluated models slightly, with fluctuations within 5\%. However, these changes do not affect the relative performance rankings of the evaluated LVLMs. Across all distractor sources, the models consistently maintain the same performance order as observed in the original experiments.

\begin{table}
    \centering
    \caption{Performance of LVLMs on selected tasks with enhanced insurance-related prompts. Accuracy (\%) is reported for each model. Performance changes are color-coded: \textcolor{green}{green} indicates improvement, while \textcolor{red}{red} denotes a decline.}
    \label{enhanced_prompt_results}
    \adjustbox{max width=0.48\textwidth}{
    \begin{tabular}{lccc}
        \toprule
        \textbf{Model}        & \textbf{House Damage Type Detection} & \textbf{Crop Growth Stage Detection} & \textbf{Vehicle Damage Severity Detection} \\
        \midrule
        GPT-4o         & 48.00/\textbf{57.00} (\textcolor{green}{+9})   & 32.00/\textbf{51.00} (\textcolor{green}{+19}) & 68.00/\textbf{80.00} (\textcolor{green}{+12})  \\
        GPT-4V         & 33.00/\textbf{40.00} (\textcolor{green}{+7})   & 22.00/\textbf{52.00} (\textcolor{green}{+30}) & 68.00/\textbf{77.00} (\textcolor{green}{+9})   \\
        Gemini 1.5 Flash   & 33.00/\textbf{47.00} (\textcolor{green}{+14})   & 28.00/\textbf{57.00} (\textcolor{green}{+29}) & 68.00/\textbf{68.00} (-)  \\
        Qwen-VL-Max    & 27.00/\textbf{42.00} (\textcolor{green}{+15})  & 30.00/\textbf{58.00} (\textcolor{green}{+28}) & 72.00/\textbf{61.00} (\textcolor{red}{-11})  \\
        Qwen-VL-Plus   & 35.00/\textbf{38.00} (\textcolor{green}{+3})   & 22.00/\textbf{60.00} (\textcolor{green}{+38}) & 68.00/\textbf{58.00} (\textcolor{red}{-10})  \\
        \bottomrule        
    \end{tabular}}
\end{table}

\begin{table}[t]
\centering
\caption{Results of different distractor generation models.}
\label{tab:different_distractor}
\vspace{-3mm}
\resizebox{0.45\textwidth}{!}{%
\begin{tabular}{cccccccccc}
\toprule
\textbf{Task} &
\multicolumn{3}{c}{\textbf{GPT4o}} &
\multicolumn{3}{c}{\textbf{QwenVLMax}} &
\multicolumn{3}{c}{\textbf{Gemini-1.5 -Flash}} \\
\cmidrule(lr){2-4} \cmidrule(lr){5-7} \cmidrule(lr){8-10}
 & origin & claude & deepseek & origin & claude & deepseek & origin & claude & deepseek \\
\midrule
Driving behavior detection & 88.60 & 85.40 & 86.60 & 74.80 & 75.40 & 76.00 & 79.20 & 80.60 & 78.60 \\
Medical image recognition & 49.00 & 50.20 & 46.60 & 44.00 & 41.40 & 40.60 & 46.40 & 43.20 & 44.60 \\
Vehicle appearance recognition & 98.20 & 93.00 & 96.00 & 98.20 & 92.60 & 95.20 & 96.80 & 91.20 & 93.40 \\
\bottomrule
\end{tabular}
}
\end{table}

\section{Discussions and conclusions}\label{sec:conclusion}

In this paper, we introduce INS-MMBench, a multimodal benchmark tailored for the insurance domain. To the best of our knowledge, this is the first initiative to systematically review multimodal tasks in this sector and to establish an insurance domain-specific benchmark. INS-MMBench comprises 12,052 images and 10,372 questions, covering 4 types of insurance, 5 scenario tasks, 12 meta-tasks, and 22 fundamental tasks, providing a comprehensive framework for assessing the applicability of LVLMs in the insurance domain. Additionally, we evaluate 11 leading LVLMs and provide a detailed analysis of the results, offering an initial exploration into the feasibility of employing LVLMs in insurance applications. Our benchmark and findings aim to guide future research and promote both interdisciplinary collaboration and practical implementation within the industry.

However, this study has some limitations. One constraint is the lack of open-source image datasets specific to the insurance domain, primarily due to privacy concerns. The image data used in this study, sourced from publicly available datasets, has been rigorously curated to align as closely as possible with real-world insurance scenarios. Nevertheless, since these images are not from actual insurance cases, there remains an inherent risk of discrepancy. This highlights the need for collaborative efforts between the insurance industry and the research community to develop dedicated open-source image datasets for this domain. Moreover, while current LVLMs perform reasonably well on basic perception tasks, they still face challenges in tasks requiring complex reasoning and specialized insurance knowledge. Future work may explore fine-tuning LVLMs on domain-specific data or integrating Retrieval-Augmented Generation (RAG), which could enable models to incorporate external knowledge and enhance their reasoning abilities, thereby improving performance in complex insurance-related scenarios.


{
    \small
    \bibliographystyle{ieeenat_fullname}
    \bibliography{custom}

\begin{thebibliography}{88}
\providecommand{\natexlab}[1]{#1}
\providecommand{\url}[1]{\texttt{#1}}
\expandafter\ifx\csname urlstyle\endcsname\relax
  \providecommand{\doi}[1]{doi: #1}\else
  \providecommand{\doi}{doi: \begingroup \urlstyle{rm}\Url}\fi

\bibitem[Abacha et~al.(2019)Abacha, Hasan, Datla, Liu, Demner-Fushman, and M{\"u}ller]{abacha2019vqa}
Asma~Ben Abacha, Sadid~A Hasan, Vivek~V Datla, Joey Liu, Dina Demner-Fushman, and Henning M{\"u}ller.
\newblock Vqa-med: Overview of the medical visual question answering task at imageclef 2019.
\newblock \emph{CLEF (working notes)}, 2\penalty0 (6), 2019.

\bibitem[Achiam et~al.(2023)Achiam, Adler, Agarwal, Ahmad, Akkaya, Aleman, Almeida, Altenschmidt, Altman, Anadkat, et~al.]{achiam2023gpt}
Josh Achiam, Steven Adler, Sandhini Agarwal, Lama Ahmad, Ilge Akkaya, Florencia~Leoni Aleman, Diogo Almeida, Janko Altenschmidt, Sam Altman, Shyamal Anadkat, et~al.
\newblock Gpt-4 technical report.
\newblock \emph{arXiv preprint arXiv:2303.08774}, 2023.

\bibitem[Agyemang(2021)]{damage-level_dataset}
Isaac Agyemang.
\newblock Damage level dataset.
\newblock \url{ https://universe.roboflow.com/isaac-agyemang/damage-level }, 2021.
\newblock visited on 2024-05-28.

\bibitem[Agyemang(2022)]{damage-type_dataset}
Isaac Agyemang.
\newblock Damage type dataset.
\newblock \url{ https://universe.roboflow.com/isaac-agyemang/damage-type }, 2022.
\newblock visited on 2024-05-28.

\bibitem[AMAN2000JAISWAL(2021)]{Agriculturecropimages}
AMAN2000JAISWAL.
\newblock Agriculture crop images.
\newblock \url{https://www.kaggle.com/datasets/aman2000jaiswal/agriculture-crop-images}, 2021.
\newblock visited on 2024-05-21.

\bibitem[ansonlau1325@gmail.com(2022)]{car-crash-severity-detection_dataset}
ansonlau1325@gmail.com.
\newblock Car crash severity detection dataset.
\newblock \url{ https://universe.roboflow.com/ansonlau1325-gmail-com/car-crash-severity-detection }, 2022.
\newblock visited on 2024-05-28.

\bibitem[Bai et~al.(2023)Bai, Bai, Yang, Wang, Tan, Wang, Lin, Zhou, and Zhou]{bai2023qwen}
Jinze Bai, Shuai Bai, Shusheng Yang, Shijie Wang, Sinan Tan, Peng Wang, Junyang Lin, Chang Zhou, and Jingren Zhou.
\newblock Qwen-vl: A frontier large vision-language model with versatile abilities.
\newblock \emph{arXiv preprint arXiv:2308.12966}, 2023.

\bibitem[Capstone2(2022)]{damages-svll3_dataset}
Capstone2.
\newblock Damages dataset.
\newblock \url{ https://universe.roboflow.com/capstone2/damages-svll3 }, 2022.
\newblock visited on 2024-05-28.

\bibitem[Chang et~al.(2024)Chang, Wang, Wang, Wu, Yang, Zhu, Chen, Yi, Wang, Wang, et~al.]{chang2024survey}
Yupeng Chang, Xu Wang, Jindong Wang, Yuan Wu, Linyi Yang, Kaijie Zhu, Hao Chen, Xiaoyuan Yi, Cunxiang Wang, Yidong Wang, et~al.
\newblock A survey on evaluation of large language models.
\newblock \emph{ACM Transactions on Intelligent Systems and Technology}, 15\penalty0 (3):\penalty0 1--45, 2024.

\bibitem[Chen et~al.(2024{\natexlab{a}})Chen, Zhou, Hua, Loh, Chen, Li, Zhu, and Liang]{chen2024fintextqa}
Jian Chen, Peilin Zhou, Yining Hua, Yingxin Loh, Kehui Chen, Ziyuan Li, Bing Zhu, and Junwei Liang.
\newblock Fintextqa: A dataset for long-form financial question answering.
\newblock \emph{arXiv preprint arXiv:2405.09980}, 2024{\natexlab{a}}.

\bibitem[Chen et~al.(2024{\natexlab{b}})Chen, Li, Dong, Zhang, Zang, Chen, Duan, Wang, Qiao, Lin, et~al.]{chen2024we}
Lin Chen, Jinsong Li, Xiaoyi Dong, Pan Zhang, Yuhang Zang, Zehui Chen, Haodong Duan, Jiaqi Wang, Yu Qiao, Dahua Lin, et~al.
\newblock Are we on the right way for evaluating large vision-language models?
\newblock \emph{arXiv preprint arXiv:2403.20330}, 2024{\natexlab{b}}.

\bibitem[Chen et~al.(2023)Chen, Wu, Wang, Su, Chen, Xing, Muyan, Zhang, Zhu, Lu, et~al.]{chen2023internvl}
Zhe Chen, Jiannan Wu, Wenhai Wang, Weijie Su, Guo Chen, Sen Xing, Zhong Muyan, Qinglong Zhang, Xizhou Zhu, Lewei Lu, et~al.
\newblock Internvl: Scaling up vision foundation models and aligning for generic visual-linguistic tasks.
\newblock \emph{arXiv preprint arXiv:2312.14238}, 2023.

\bibitem[Chew et~al.(2020)Chew, Rineer, Beach, O’Neil, Ujeneza, Lapidus, Miano, Hegarty-Craver, Polly, and Temple]{chew2020deep}
Robert Chew, Jay Rineer, Robert Beach, Maggie O’Neil, Noel Ujeneza, Daniel Lapidus, Thomas Miano, Meghan Hegarty-Craver, Jason Polly, and Dorota~S Temple.
\newblock Deep neural networks and transfer learning for food crop identification in uav images.
\newblock \emph{Drones}, 4\penalty0 (1):\penalty0 7, 2020.

\bibitem[Chiu et~al.(2020)Chiu, Xu, Wei, Huang, Schwing, Brunner, Khachatrian, Karapetyan, Dozier, Rose, et~al.]{chiu2020agriculture}
Mang~Tik Chiu, Xingqian Xu, Yunchao Wei, Zilong Huang, Alexander~G Schwing, Robert Brunner, Hrant Khachatrian, Hovnatan Karapetyan, Ivan Dozier, Greg Rose, et~al.
\newblock Agriculture-vision: A large aerial image database for agricultural pattern analysis.
\newblock In \emph{Proceedings of the IEEE/CVF Conference on Computer Vision and Pattern Recognition}, pages 2828--2838, 2020.

\bibitem[College(2023)]{fire-detection-cta61_dataset}
College.
\newblock fire detection dataset.
\newblock \url{ https://universe.roboflow.com/college-pbetq/fire-detection-cta61 }, 2023.
\newblock visited on 2024-05-28.

\bibitem[computer vision(2022)]{worker-safety_dataset}
computer vision.
\newblock Worker-safety dataset.
\newblock \url{ https://universe.roboflow.com/computer-vision/worker-safety }, 2022.
\newblock visited on 2024-05-28.

\bibitem[Dashboarddataset(2024)]{dataset_dashboard__dataset}
Dashboarddataset.
\newblock dataset dashboard dataset.
\newblock \url{ https://universe.roboflow.com/dashboarddataset/dataset_dashboard_ }, 2024.
\newblock visited on 2024-05-28.

\bibitem[Demner-Fushman et~al.(2016)Demner-Fushman, Kohli, Rosenman, Shooshan, Rodriguez, Antani, Thoma, and McDonald]{demner2016preparing}
Dina Demner-Fushman, Marc~D Kohli, Marc~B Rosenman, Sonya~E Shooshan, Laritza Rodriguez, Sameer Antani, George~R Thoma, and Clement~J McDonald.
\newblock Preparing a collection of radiology examinations for distribution and retrieval.
\newblock \emph{Journal of the American Medical Informatics Association}, 23\penalty0 (2):\penalty0 304--310, 2016.

\bibitem[Deshmukh et~al.(2023)Deshmukh, Elizalde, Singh, and Wang]{deshmukh2023pengi}
Soham Deshmukh, Benjamin Elizalde, Rita Singh, and Huaming Wang.
\newblock Pengi: An audio language model for audio tasks.
\newblock \emph{Advances in Neural Information Processing Systems}, 36:\penalty0 18090--18108, 2023.

\bibitem[Dewangan et~al.(2023)Dewangan, Choudhary, Chandhok, Priyadarshan, Jain, Singh, Srivastava, Jatavallabhula, and Krishna]{dewangan2023talk2bev}
Vikrant Dewangan, Tushar Choudhary, Shivam Chandhok, Shubham Priyadarshan, Anushka Jain, Arun~K Singh, Siddharth Srivastava, Krishna~Murthy Jatavallabhula, and K~Madhava Krishna.
\newblock Talk2bev: Language-enhanced bird's-eye view maps for autonomous driving.
\newblock \emph{arXiv preprint arXiv:2310.02251}, 2023.

\bibitem[Driver et~al.(2018)Driver, Brimble, Freudenberg, and Hunt]{driver2018insurance}
Tania Driver, Mark Brimble, Brett Freudenberg, and Katherine Hunt.
\newblock Insurance literacy in australia: Not knowing the value of personal insurance.
\newblock \emph{Financial Planning Research Journal}, 4\penalty0 (1):\penalty0 53--75, 2018.

\bibitem[Duan et~al.(2024)Duan, Yang, Qiao, Fang, Chen, Liu, Dong, Zang, Zhang, Wang, Lin, and Chen]{duan2024vlmevalkit}
Haodong Duan, Junming Yang, Yuxuan Qiao, Xinyu Fang, Lin Chen, Yuan Liu, Xiaoyi Dong, Yuhang Zang, Pan Zhang, Jiaqi Wang, Dahua Lin, and Kai Chen.
\newblock Vlmevalkit: An open-source toolkit for evaluating large multi-modality models, 2024.

\bibitem[DUTTA(2023)]{WheatGrowthStageChallenge}
GAURAV DUTTA.
\newblock Wheat growth stage challenge.
\newblock \url{https://www.kaggle.com/datasets/gauravduttakiit/wheat-growth-stage-challenge}, 2023.
\newblock visited on 2024-05-21.

\bibitem[Eling and Lehmann(2018)]{eling2018impact}
Martin Eling and Martin Lehmann.
\newblock The impact of digitalization on the insurance value chain and the insurability of risks.
\newblock \emph{The Geneva papers on risk and insurance-issues and practice}, 43:\penalty0 359--396, 2018.

\bibitem[Eling et~al.(2022)Eling, Nuessle, and Staubli]{eling2022impact}
Martin Eling, Davide Nuessle, and Julian Staubli.
\newblock The impact of artificial intelligence along the insurance value chain and on the insurability of risks.
\newblock \emph{The Geneva Papers on Risk and Insurance-Issues and Practice}, 47\penalty0 (2):\penalty0 205--241, 2022.

\bibitem[Ezzouhri et~al.(2021)Ezzouhri, Charouh, Ghogho, and Guennoun]{ezzouhri2021robust}
Amal Ezzouhri, Zakaria Charouh, Mounir Ghogho, and Zouhair Guennoun.
\newblock Robust deep learning-based driver distraction detection and classification.
\newblock \emph{IEEE Access}, 9:\penalty0 168080--168092, 2021.

\bibitem[f-rid nagiyev(2023)]{tuning-car-detection_dataset}
f-rid nagiyev.
\newblock Tuning car detection dataset.
\newblock \url{ https://universe.roboflow.com/f-rid-nagiyev/tuning-car-detection }, 2023.
\newblock visited on 2024-05-28.

\bibitem[Fang et~al.(2021)Fang, Yan, Qiao, Xue, and Yu]{fang2021dada}
Jianwu Fang, Dingxin Yan, Jiahuan Qiao, Jianru Xue, and Hongkai Yu.
\newblock Dada: Driver attention prediction in driving accident scenarios.
\newblock \emph{IEEE transactions on intelligent transportation systems}, 23\penalty0 (6):\penalty0 4959--4971, 2021.

\bibitem[Fernando et~al.(2022)Fernando, Kumarage, Thiyaganathan, Hillary, and Abeywardhana]{fernando2022automated}
Nisaja Fernando, Abimani Kumarage, Vithyashagar Thiyaganathan, Radesh Hillary, and Lakmini Abeywardhana.
\newblock Automated vehicle insurance claims processing using computer vision, natural language processing.
\newblock In \emph{2022 22nd International Conference on Advances in ICT for Emerging Regions (ICTer)}, pages 124--129. IEEE, 2022.

\bibitem[Fu et~al.(2023)Fu, Zhang, Lin, Wang, Gao, Luo, Huang, Zhang, Qiu, Ye, et~al.]{fu2023challenger}
Chaoyou Fu, Renrui Zhang, Haojia Lin, Zihan Wang, Timin Gao, Yongdong Luo, Yubo Huang, Zhengye Zhang, Longtian Qiu, Gaoxiang Ye, et~al.
\newblock A challenger to gpt-4v? early explorations of gemini in visual expertise.
\newblock \emph{arXiv preprint arXiv:2312.12436}, 2023.

\bibitem[Google(2024)]{gemini}
Google.
\newblock Gemini pro.
\newblock \url{https://deepmind.google/technologies/gemini/pro/}, 2024.
\newblock Accessed: 2024-05-23.

\bibitem[Gupta et~al.(2019)Gupta, Goodman, Patel, Hosfelt, Sajeev, Heim, Doshi, Lucas, Choset, and Gaston]{gupta2019creating}
Ritwik Gupta, Bryce Goodman, Nirav Patel, Ricky Hosfelt, Sandra Sajeev, Eric Heim, Jigar Doshi, Keane Lucas, Howie Choset, and Matthew Gaston.
\newblock Creating xbd: A dataset for assessing building damage from satellite imagery.
\newblock In \emph{Proceedings of the IEEE/CVF conference on computer vision and pattern recognition workshops}, pages 10--17, 2019.

\bibitem[{HackerEarth}(2020)]{HackerEarthInsurance}
{HackerEarth}.
\newblock Hackerearth machine learning challenge: Vehicle insurance claim, 2020.
\newblock Accessed: Jan 6, 2025.

\bibitem[Hu et~al.(2024)Hu, Li, Lu, Shao, He, Qiao, and Luo]{hu2024omnimedvqa}
Yutao Hu, Tianbin Li, Quanfeng Lu, Wenqi Shao, Junjun He, Yu Qiao, and Ping Luo.
\newblock Omnimedvqa: A new large-scale comprehensive evaluation benchmark for medical lvlm.
\newblock \emph{arXiv preprint arXiv:2402.09181}, 2024.

\bibitem[Huang et~al.(2022)Huang, Gu, Hou, Wu, Wang, Yu, and Han]{huang2022large}
Jiaxin Huang, Shixiang~Shane Gu, Le Hou, Yuexin Wu, Xuezhi Wang, Hongkun Yu, and Jiawei Han.
\newblock Large language models can self-improve.
\newblock \emph{arXiv preprint arXiv:2210.11610}, 2022.

\bibitem[KANDAGATLA(2022)]{fall_detection_detection}
UTTEJ~KUMAR KANDAGATLA.
\newblock Fall detection dataset.
\newblock \url{https://www.kaggle.com/datasets/uttejkumarkandagatla/fall-detection-dataset}, 2022.
\newblock visited on 2024-05-20.

\bibitem[Kasneci et~al.(2023)Kasneci, Se{\ss}ler, K{\"u}chemann, Bannert, Dementieva, Fischer, Gasser, Groh, G{\"u}nnemann, H{\"u}llermeier, et~al.]{kasneci2023chatgpt}
Enkelejda Kasneci, Kathrin Se{\ss}ler, Stefan K{\"u}chemann, Maria Bannert, Daryna Dementieva, Frank Fischer, Urs Gasser, Georg Groh, Stephan G{\"u}nnemann, Eyke H{\"u}llermeier, et~al.
\newblock Chatgpt for good? on opportunities and challenges of large language models for education.
\newblock \emph{Learning and individual differences}, 103:\penalty0 102274, 2023.

\bibitem[Krause et~al.(2013)Krause, Stark, Deng, and Fei-Fei]{krause20133d}
Jonathan Krause, Michael Stark, Jia Deng, and Li Fei-Fei.
\newblock 3d object representations for fine-grained categorization.
\newblock In \emph{Proceedings of the IEEE international conference on computer vision workshops}, pages 554--561, 2013.

\bibitem[Li et~al.(2023{\natexlab{a}})Li, Ge, Ge, Wang, Wang, Zhang, and Shan]{li2023seed2}
Bohao Li, Yuying Ge, Yixiao Ge, Guangzhi Wang, Rui Wang, Ruimao Zhang, and Ying Shan.
\newblock Seed-bench-2: Benchmarking multimodal large language models.
\newblock \emph{arXiv preprint arXiv:2311.17092}, 2023{\natexlab{a}}.

\bibitem[Li et~al.(2023{\natexlab{b}})Li, Wang, Wang, Ge, Ge, and Shan]{li2023seed1}
Bohao Li, Rui Wang, Guangzhi Wang, Yuying Ge, Yixiao Ge, and Ying Shan.
\newblock Seed-bench: Benchmarking multimodal llms with generative comprehension.
\newblock \emph{arXiv preprint arXiv:2307.16125}, 2023{\natexlab{b}}.

\bibitem[Li et~al.(2024{\natexlab{a}})Li, Ge, Chen, Ge, Zhang, and Shan]{li2024seed}
Bohao Li, Yuying Ge, Yi Chen, Yixiao Ge, Ruimao Zhang, and Ying Shan.
\newblock Seed-bench-2-plus: Benchmarking multimodal large language models with text-rich visual comprehension.
\newblock \emph{arXiv preprint arXiv:2404.16790}, 2024{\natexlab{a}}.

\bibitem[Li et~al.(2023{\natexlab{c}})Li, Li, Savarese, and Hoi]{li2023blip}
Junnan Li, Dongxu Li, Silvio Savarese, and Steven Hoi.
\newblock Blip-2: Bootstrapping language-image pre-training with frozen image encoders and large language models.
\newblock In \emph{International conference on machine learning}, pages 19730--19742. PMLR, 2023{\natexlab{c}}.

\bibitem[Li et~al.(2023{\natexlab{d}})Li, Wang, He, Li, Wang, Liu, Wang, Xu, Chen, Luo, et~al.]{li2023mvbench}
Kunchang Li, Yali Wang, Yinan He, Yizhuo Li, Yi Wang, Yi Liu, Zun Wang, Jilan Xu, Guo Chen, Ping Luo, et~al.
\newblock Mvbench: A comprehensive multi-modal video understanding benchmark.
\newblock \emph{arXiv preprint arXiv:2311.17005}, 2023{\natexlab{d}}.

\bibitem[Li et~al.(2018)Li, Shen, and Dong]{li2018anti}
Pei Li, Bingyu Shen, and Weishan Dong.
\newblock An anti-fraud system for car insurance claim based on visual evidence.
\newblock \emph{arXiv preprint arXiv:1804.11207}, 2018.

\bibitem[Li et~al.(2023{\natexlab{e}})Li, Wang, Hu, Chen, Zhong, Lyu, and Zhang]{li2023comprehensive}
Yunxin Li, Longyue Wang, Baotian Hu, Xinyu Chen, Wanqi Zhong, Chenyang Lyu, and Min Zhang.
\newblock A comprehensive evaluation of gpt-4v on knowledge-intensive visual question answering.
\newblock \emph{arXiv preprint arXiv:2311.07536}, 2023{\natexlab{e}}.

\bibitem[Li et~al.(2023{\natexlab{f}})Li, Wang, Ding, and Chen]{li2023large}
Yinheng Li, Shaofei Wang, Han Ding, and Hang Chen.
\newblock Large language models in finance: A survey.
\newblock In \emph{Proceedings of the Fourth ACM International Conference on AI in Finance}, pages 374--382, 2023{\natexlab{f}}.

\bibitem[Li et~al.(2024{\natexlab{b}})Li, Zhang, Chen, Liu, Li, Gao, Hong, Tian, Zhao, Li, et~al.]{li2024automated}
Yanze Li, Wenhua Zhang, Kai Chen, Yanxin Liu, Pengxiang Li, Ruiyuan Gao, Lanqing Hong, Meng Tian, Xinhai Zhao, Zhenguo Li, et~al.
\newblock Automated evaluation of large vision-language models on self-driving corner cases.
\newblock \emph{arXiv preprint arXiv:2404.10595}, 2024{\natexlab{b}}.

\bibitem[Lin et~al.(2024)Lin, Lyu, Luo, and Xu]{lin2024harnessing}
Chenwei Lin, Hanjia Lyu, Jiebo Luo, and Xian Xu.
\newblock Harnessing gpt-4v (ision) for insurance: A preliminary exploration.
\newblock \emph{arXiv preprint arXiv:2404.09690}, 2024.

\bibitem[Liu et~al.(2023{\natexlab{a}})Liu, Lin, Li, Wang, Yacoob, and Wang]{liu2023mitigating}
Fuxiao Liu, Kevin Lin, Linjie Li, Jianfeng Wang, Yaser Yacoob, and Lijuan Wang.
\newblock Mitigating hallucination in large multi-modal models via robust instruction tuning.
\newblock In \emph{The Twelfth International Conference on Learning Representations}, 2023{\natexlab{a}}.

\bibitem[Liu et~al.(2023{\natexlab{b}})Liu, Wang, Yao, Chen, Song, Cho, Yacoob, and Yu]{liu2023mmc}
Fuxiao Liu, Xiaoyang Wang, Wenlin Yao, Jianshu Chen, Kaiqiang Song, Sangwoo Cho, Yaser Yacoob, and Dong Yu.
\newblock Mmc: Advancing multimodal chart understanding with large-scale instruction tuning.
\newblock \emph{arXiv preprint arXiv:2311.10774}, 2023{\natexlab{b}}.

\bibitem[Liu et~al.(2024)Liu, Li, Wu, and Lee]{liu2024visual}
Haotian Liu, Chunyuan Li, Qingyang Wu, and Yong~Jae Lee.
\newblock Visual instruction tuning.
\newblock \emph{Advances in neural information processing systems}, 36, 2024.

\bibitem[Liu et~al.(2023{\natexlab{c}})Liu, Duan, Zhang, Li, Zhang, Zhao, Yuan, Wang, He, Liu, et~al.]{liu2023mmbench}
Yuan Liu, Haodong Duan, Yuanhan Zhang, Bo Li, Songyang Zhang, Wangbo Zhao, Yike Yuan, Jiaqi Wang, Conghui He, Ziwei Liu, et~al.
\newblock Mmbench: Is your multi-modal model an all-around player?
\newblock \emph{arXiv preprint arXiv:2307.06281}, 2023{\natexlab{c}}.

\bibitem[Lu et~al.(2023)Lu, Bansal, Xia, Liu, Li, Hajishirzi, Cheng, Chang, Galley, and Gao]{lu2023mathvista}
Pan Lu, Hritik Bansal, Tony Xia, Jiacheng Liu, Chunyuan Li, Hannaneh Hajishirzi, Hao Cheng, Kai-Wei Chang, Michel Galley, and Jianfeng Gao.
\newblock Mathvista: Evaluating mathematical reasoning of foundation models in visual contexts.
\newblock \emph{arXiv preprint arXiv:2310.02255}, 2023.

\bibitem[Lyu et~al.(2025)Lyu, Huang, Zhang, Yu, Mou, Pan, Yang, Wei, and Luo]{lyu2025gpt}
Hanjia Lyu, Jinfa Huang, Daoan Zhang, Yongsheng Yu, Xinyi Mou, Jinsheng Pan, Zhengyuan Yang, Zhongyu Wei, and Jiebo Luo.
\newblock Gpt-4v(ision) as a social media analysis engine.
\newblock \emph{ACM Trans. Intell. Syst. Technol.}, 16\penalty0 (3), 2025.

\bibitem[Mallios et~al.(2023)Mallios, Xiaofei, McLaughlin, Del~Rincon, Galbraith, and Garland]{mallios2023vehicle}
Dimitrios Mallios, Li Xiaofei, Niall McLaughlin, Jesus~Martinez Del~Rincon, Clare Galbraith, and Rory Garland.
\newblock Vehicle damage severity estimation for insurance operations using in-the-wild mobile images.
\newblock \emph{IEEE Access}, 2023.

\bibitem[Mouheb et~al.(2021)Mouheb, Y{\"u}rekli, and Y{\i}lmazel]{mouheb2021trodo}
Kaouther Mouheb, Ali Y{\"u}rekli, and Burcu Y{\i}lmazel.
\newblock Trodo: A public vehicle odometers dataset for computer vision.
\newblock \emph{Data in Brief}, 38:\penalty0 107321, 2021.

\bibitem[OpenAI(2024)]{GPT4o}
OpenAI.
\newblock Hello gpt-4o.
\newblock \url{https://openai.com/index/hello-gpt-4o/}, 2024.
\newblock Accessed: 2024-05-23.

\bibitem[Project(2024)]{blood-pressure-monitor-display_dataset}
Final Project.
\newblock blood-pressure-monitor-display dataset.
\newblock \url{ https://universe.roboflow.com/final-project-cwtfb/blood-pressure-monitor-display }, 2024.
\newblock visited on 2024-05-28.

\bibitem[Roberts et~al.(2023)Roberts, L{\"u}ddecke, Sheikh, Han, and Albanie]{roberts2023charting}
Jonathan Roberts, Timo L{\"u}ddecke, Rehan Sheikh, Kai Han, and Samuel Albanie.
\newblock Charting new territories: Exploring the geographic and geospatial capabilities of multimodal llms.
\newblock \emph{arXiv preprint arXiv:2311.14656}, 2023.

\bibitem[Roberts et~al.(2024)Roberts, Han, Houlsby, and Albanie]{roberts2024scifibench}
Jonathan Roberts, Kai Han, Neil Houlsby, and Samuel Albanie.
\newblock Scifibench: Benchmarking large multimodal models for scientific figure interpretation.
\newblock \emph{arXiv preprint arXiv:2405.08807}, 2024.

\bibitem[Sahni et~al.(2020)Sahni, Mittal, Kidwai, Tiwari, and Khandelwal]{sahni2020insurance}
Srishti Sahni, Anmol Mittal, Farzil Kidwai, Ajay Tiwari, and Kanak Khandelwal.
\newblock Insurance fraud identification using computer vision and iot: a study of field fires.
\newblock \emph{Procedia Computer Science}, 173:\penalty0 56--63, 2020.

\bibitem[Shen et~al.(2023)Shen, Heacock, Elias, Hentel, Reig, Shih, and Moy]{shen2023chatgpt}
Yiqiu Shen, Laura Heacock, Jonathan Elias, Keith~D Hentel, Beatriu Reig, George Shih, and Linda Moy.
\newblock Chatgpt and other large language models are double-edged swords, 2023.

\bibitem[Sindhu(2022)]{car_dent_scratch_detection-1_dataset}
Sindhu.
\newblock Car dent scratch detection(1) dataset.
\newblock \url{ https://universe.roboflow.com/sindhu/car_dent_scratch_detection-1 }, 2022.
\newblock visited on 2024-05-28.

\bibitem[Team(2024)]{QwenVL}
Qwen Team.
\newblock Introducing qwen-vl.
\newblock \url{https://qwenlm.github.io/blog/qwen-vl/}, 2024.
\newblock Accessed: 2024-05-23.

\bibitem[Vélez et~al.(2023)Vélez, Ariza-Sentís, and Valente]{velez_2023_10362568}
Sergio Vélez, Mar Ariza-Sentís, and João Valente.
\newblock Precision viticulture dataset for detailed vineyard mapping composed of geotagged smartphone ground images, phytosanitary status, uav orthomosaics, 3d point clouds, and rtk gnss data - northern spain, july 2022, 2023.

\bibitem[Wang et~al.(2024{\natexlab{a}})Wang, Bai, Nah, Wang, Zhang, Chen, Wu, Islam, Liu, and Ren]{wang2024surgical}
Guankun Wang, Long Bai, Wan~Jun Nah, Jie Wang, Zhaoxi Zhang, Zhen Chen, Jinlin Wu, Mobarakol Islam, Hongbin Liu, and Hongliang Ren.
\newblock Surgical-lvlm: Learning to adapt large vision-language model for grounded visual question answering in robotic surgery.
\newblock \emph{arXiv preprint arXiv:2405.10948}, 2024{\natexlab{a}}.

\bibitem[Wang et~al.(2024{\natexlab{b}})Wang, Pan, Shi, Lu, Zhan, and Li]{wang2024measuring}
Ke Wang, Junting Pan, Weikang Shi, Zimu Lu, Mingjie Zhan, and Hongsheng Li.
\newblock Measuring multimodal mathematical reasoning with math-vision dataset.
\newblock \emph{arXiv preprint arXiv:2402.14804}, 2024{\natexlab{b}}.

\bibitem[Wang et~al.(2024{\natexlab{c}})Wang, Chen, Chen, Wu, Zhu, Zeng, Luo, Lu, Zhou, Qiao, et~al.]{wang2024visionllm}
Wenhai Wang, Zhe Chen, Xiaokang Chen, Jiannan Wu, Xizhou Zhu, Gang Zeng, Ping Luo, Tong Lu, Jie Zhou, Yu Qiao, et~al.
\newblock Visionllm: Large language model is also an open-ended decoder for vision-centric tasks.
\newblock \emph{Advances in Neural Information Processing Systems}, 36, 2024{\natexlab{c}}.

\bibitem[Wang et~al.(2023)Wang, Li, and Wu]{wang2023cardd}
Xinkuang Wang, Wenjing Li, and Zhongcheng Wu.
\newblock Cardd: A new dataset for vision-based car damage detection.
\newblock \emph{IEEE Transactions on Intelligent Transportation Systems}, 2023.

\bibitem[Weedige et~al.(2019)Weedige, Ouyang, Gao, and Liu]{weedige2019decision}
Sampath~Sanjeewa Weedige, Hongbing Ouyang, Yao Gao, and Yaqing Liu.
\newblock Decision making in personal insurance: Impact of insurance literacy.
\newblock \emph{Sustainability}, 11\penalty0 (23):\penalty0 6795, 2019.

\bibitem[Wei et~al.(2022)Wei, Tay, Bommasani, Raffel, Zoph, Borgeaud, Yogatama, Bosma, Zhou, Metzler, et~al.]{wei2022emergent}
Jason Wei, Yi Tay, Rishi Bommasani, Colin Raffel, Barret Zoph, Sebastian Borgeaud, Dani Yogatama, Maarten Bosma, Denny Zhou, Donald Metzler, et~al.
\newblock Emergent abilities of large language models.
\newblock \emph{arXiv preprint arXiv:2206.07682}, 2022.

\bibitem[workspace(2023)]{mjdfodf-qmbuf_dataset}
workspace.
\newblock mjdfodf-qmbuf dataset.
\newblock \url{ https://universe.roboflow.com/workspace-luixd/mjdfodf-qmbuf }, 2023.
\newblock visited on 2024-05-28.

\bibitem[Wu et~al.(2023)Wu, Gan, Chen, Wan, and Philip]{wu2023multimodal}
Jiayang Wu, Wensheng Gan, Zefeng Chen, Shicheng Wan, and S~Yu Philip.
\newblock Multimodal large language models: A survey.
\newblock In \emph{2023 IEEE International Conference on Big Data (BigData)}, pages 2247--2256. IEEE, 2023.

\bibitem[Xu et~al.(2023)Xu, Shao, Zhang, Gao, Liu, Lei, Meng, Huang, Qiao, and Luo]{xu2023lvlm}
Peng Xu, Wenqi Shao, Kaipeng Zhang, Peng Gao, Shuo Liu, Meng Lei, Fanqing Meng, Siyuan Huang, Yu Qiao, and Ping Luo.
\newblock Lvlm-ehub: A comprehensive evaluation benchmark for large vision-language models.
\newblock \emph{arXiv preprint arXiv:2306.09265}, 2023.

\bibitem[Xu et~al.(2021)Xu, Wang, Shou, Ngo, Sadick, and Wang]{xu2021computer}
Shuyuan Xu, Jun Wang, Wenchi Shou, Tuan Ngo, Abdul-Manan Sadick, and Xiangyu Wang.
\newblock Computer vision techniques in construction: a critical review.
\newblock \emph{Archives of Computational Methods in Engineering}, 28:\penalty0 3383--3397, 2021.

\bibitem[Xu et~al.(2018)Xu, Yang, Meng, Lu, Huang, Ying, and Huang]{xu2018towards}
Zhenbo Xu, Wei Yang, Ajin Meng, Nanxue Lu, Huan Huang, Changchun Ying, and Liusheng Huang.
\newblock Towards end-to-end license plate detection and recognition: A large dataset and baseline.
\newblock In \emph{Proceedings of the European conference on computer vision (ECCV)}, pages 255--271, 2018.

\bibitem[Yang et~al.(2023)Yang, Li, Lin, Wang, Lin, Liu, and Wang]{yang2023dawn}
Zhengyuan Yang, Linjie Li, Kevin Lin, Jianfeng Wang, Chung-Ching Lin, Zicheng Liu, and Lijuan Wang.
\newblock The dawn of lmms: Preliminary explorations with gpt-4v (ision).
\newblock \emph{arXiv preprint arXiv:2309.17421}, 9\penalty0 (1):\penalty0 1, 2023.

\bibitem[Ye et~al.(2023)Ye, Xu, Xu, Ye, Yan, Zhou, Wang, Hu, Shi, Shi, et~al.]{ye2023mplug}
Qinghao Ye, Haiyang Xu, Guohai Xu, Jiabo Ye, Ming Yan, Yiyang Zhou, Junyang Wang, Anwen Hu, Pengcheng Shi, Yaya Shi, et~al.
\newblock mplug-owl: Modularization empowers large language models with multimodality.
\newblock \emph{arXiv preprint arXiv:2304.14178}, 2023.

\bibitem[Yin et~al.(2023)Yin, Fu, Zhao, Li, Sun, Xu, and Chen]{yin2023survey}
Shukang Yin, Chaoyou Fu, Sirui Zhao, Ke Li, Xing Sun, Tong Xu, and Enhong Chen.
\newblock A survey on multimodal large language models.
\newblock \emph{arXiv preprint arXiv:2306.13549}, 2023.

\bibitem[Ying et~al.(2024)Ying, Meng, Wang, Li, Lin, Yang, Zhang, Zhang, Lin, Liu, et~al.]{ying2024mmt}
Kaining Ying, Fanqing Meng, Jin Wang, Zhiqian Li, Han Lin, Yue Yang, Hao Zhang, Wenbo Zhang, Yuqi Lin, Shuo Liu, et~al.
\newblock Mmt-bench: A comprehensive multimodal benchmark for evaluating large vision-language models towards multitask agi.
\newblock \emph{arXiv preprint arXiv:2404.16006}, 2024.

\bibitem[Zhang et~al.(2023)Zhang, Han, Liu, Gao, Zhou, Hu, Yan, Lu, Li, and Qiao]{zhang2023llama}
Renrui Zhang, Jiaming Han, Chris Liu, Peng Gao, Aojun Zhou, Xiangfei Hu, Shilin Yan, Pan Lu, Hongsheng Li, and Yu Qiao.
\newblock Llama-adapter: Efficient fine-tuning of language models with zero-init attention.
\newblock \emph{arXiv preprint arXiv:2303.16199}, 2023.

\bibitem[Zhang et~al.(2020)Zhang, Cheng, Guo, Guo, Wang, Wang, Jiang, Wang, Xu, and Chu]{zhang2020automatic}
Wei Zhang, Yuan Cheng, Xin Guo, Qingpei Guo, Jian Wang, Qing Wang, Chen Jiang, Meng Wang, Furong Xu, and Wei Chu.
\newblock Automatic car damage assessment system: Reading and understanding videos as professional insurance inspectors.
\newblock In \emph{Proceedings of the AAAI Conference on Artificial Intelligence}, pages 13646--13647, 2020.

\bibitem[Zhang et~al.(2024{\natexlab{a}})Zhang, Aljunied, Gao, Chia, and Bing]{zhang2024m3exam}
Wenxuan Zhang, Mahani Aljunied, Chang Gao, Yew~Ken Chia, and Lidong Bing.
\newblock M3exam: A multilingual, multimodal, multilevel benchmark for examining large language models.
\newblock \emph{Advances in Neural Information Processing Systems}, 36, 2024{\natexlab{a}}.

\bibitem[Zhang et~al.(2024{\natexlab{b}})Zhang, Kuang, Mou, Lyu, Wu, Chen, Luo, Huang, and Wei]{zhang2024somelvlm}
Xinnong Zhang, Haoyu Kuang, Xinyi Mou, Hanjia Lyu, Kun Wu, Siming Chen, Jiebo Luo, Xuanjing Huang, and Zhongyu Wei.
\newblock {S}o{M}e{LVLM}: A large vision language model for social media processing.
\newblock In \emph{Findings of the Association for Computational Linguistics: ACL 2024}, pages 2366--2389, Bangkok, Thailand, 2024{\natexlab{b}}. Association for Computational Linguistics.

\bibitem[Zhang et~al.(2022)Zhang, Zhang, Li, and Smola]{zhang2022automatic}
Zhuosheng Zhang, Aston Zhang, Mu Li, and Alex Smola.
\newblock Automatic chain of thought prompting in large language models.
\newblock \emph{arXiv preprint arXiv:2210.03493}, 2022.

\bibitem[Zhao et~al.(2023{\natexlab{a}})Zhao, Zhou, Li, Tang, Wang, Hou, Min, Zhang, Zhang, Dong, et~al.]{zhao2023survey}
Wayne~Xin Zhao, Kun Zhou, Junyi Li, Tianyi Tang, Xiaolei Wang, Yupeng Hou, Yingqian Min, Beichen Zhang, Junjie Zhang, Zican Dong, et~al.
\newblock A survey of large language models.
\newblock \emph{arXiv preprint arXiv:2303.18223}, 2023{\natexlab{a}}.

\bibitem[Zhao et~al.(2023{\natexlab{b}})Zhao, Misra, Kr{\"a}henb{\"u}hl, and Girdhar]{zhao2023learning}
Yue Zhao, Ishan Misra, Philipp Kr{\"a}henb{\"u}hl, and Rohit Girdhar.
\newblock Learning video representations from large language models.
\newblock In \emph{Proceedings of the IEEE/CVF Conference on Computer Vision and Pattern Recognition}, pages 6586--6597, 2023{\natexlab{b}}.

\bibitem[Zhu et~al.(2023)Zhu, Chen, Shen, Li, and Elhoseiny]{zhu2023minigpt}
Deyao Zhu, Jun Chen, Xiaoqian Shen, Xiang Li, and Mohamed Elhoseiny.
\newblock Minigpt-4: Enhancing vision-language understanding with advanced large language models.
\newblock \emph{arXiv preprint arXiv:2304.10592}, 2023.

\end{thebibliography}
}

\end{document}